\documentclass{article}

    \PassOptionsToPackage{numbers, compress}{natbib}

\usepackage[final]{neurips_2022}

\usepackage[utf8]{inputenc} 
\usepackage[T1]{fontenc}    
\usepackage{hyperref}       
\usepackage{url}            
\usepackage{booktabs}       
\usepackage{amsfonts}       
\usepackage{nicefrac}       
\usepackage{microtype}      
\usepackage{xcolor}         

\usepackage{graphicx}
\usepackage{amsmath, amssymb, amsthm}
\usepackage{adjustbox}
\usepackage{caption}
\usepackage{esint}
\usepackage{hyperref, multicol}
\hypersetup{colorlinks=false, allcolors=blue}

\usepackage [english]{babel}
\theoremstyle{definition}

\theoremstyle{definition}

\usepackage{float}
\usepackage{verbatim}
\usepackage{euscript}
\usepackage{enumitem}
\usepackage{float}
\usepackage{tcolorbox}
\usepackage{subfig}
\usepackage{graphicx}
\usepackage{tcolorbox}
\usepackage[english]{babel}
\usepackage{natbib}
\usepackage{pgfplots}
\pgfplotsset{compat=newest}
\pgfplotsset{scaled y ticks=false}
\usepgfplotslibrary{groupplots}
\usepgfplotslibrary{dateplot}
\usepackage{tikz}

\DeclareMathOperator*{\cossim}{\textnormal{CosSim}}
\DeclareMathOperator*{\SCS}{\textnormal{SCS}}

\title{Exploring the Sharpened Cosine Similarity}

\author{%
  Skyler Wu$^{1,2,4}$, Fred Lu$^{1,2,3}$, Edward Raff$^{1,2,3}$, James Holt$^{1}$\\
  $^{1}$Laboratory for Physical Sciences, $^{2}$Booz Allen Hamilton,\\$^{3}$ University of Maryland, Baltimore County, $^{4}$Harvard University\\
  \texttt{skylerwu@college.harvard.edu}, \texttt{lu\_fred@bah.com},\\ \texttt{raff\_edward@bah.com}, \texttt{holt@lps.umd.edu} \\
}

\begin{document}

\maketitle

\begin{abstract}
    Convolutional layers have long served as the primary workhorse for image classification. Recently, an alternative to convolution was proposed using the Sharpened Cosine Similarity (SCS), which in theory may serve as a better feature detector. While multiple sources report promising results, there has not been to date a full-scale empirical analysis of neural network performance using these new layers. In our work, we explore SCS's parameter behavior and potential as a drop-in replacement for convolutions in multiple CNN architectures benchmarked on CIFAR-10. We find that while SCS may not yield significant increases in accuracy, it may learn more interpretable representations. We also find that, in some circumstances, SCS may confer a slight increase in adversarial robustness.
\end{abstract}

\section{Introduction}
For decades, convolutional layers have served as the workhorses in neural network architectures for image classification. Mathematically, a convolutional layer slides across an input image and computes the dot product between a signal $s$ and a kernel $k$ (see \cite{rohrer_2022github}). However, initial exploration and discussion by \cite{rohrer_2020} suggest that convolutional layers, while excellent image filters, may not be very good feature detectors. In addition, \cite{luo2018cosine} argue that because the dot product operation is unbounded, neural networks using convolutional layers may be vulnerable to increased model variance, over-sensitivity, and a lack of generalizability. As such, \cite{luo2018cosine} explored replacing traditional convolutional layers with cosine similarity. Mathematically, the cosine similarity (``CosSim") between $s$ and $k$ is defined as follows:
\[\cossim(s, k) = \frac{s \cdot k}{\lVert s\rVert \lVert k\rVert}.\]
Importantly, cosine similarity is bounded between $-1$ and $1$, potentially reducing the variance concerns of standard CNNs. Traditionally, cosine similarity has been used extensively in text-analysis to quantify the ``similarity" between two documents (see \cite{singhal2001modern}), among other machine learning tasks (see \cite{tan2013data}). For deep-learning networks, \cite{luo2018cosine} found that incorporating cosine similarity in VGG-type CNNs outperformed standard convolution-based alternatives using batch, weight, and layer normalization on CIFAR-10/100 and SVHN.

Building on cosine similarity, \cite{rohrer_2022github} proposed that one can ``sharpen" the standard cosine similarity to better discern whether two vectors are similar by raising the standard cosine similarity to a power $p$, while preserving the sign of the original cosine value. To prevent numerical instability arising from the signal $s$ potentially having a near-zero magnitude, \cite{rohrer_2022github} presents the following modified mathematical formulation for Sharpened Cosine Similarity (SCS), with a small positive scalar $q$ added to $||s||$:
\[\SCS(s,k) = \textnormal{sign}(s \cdot k) \Big| \frac{s \cdot k}{(\lVert s\rVert + q)\lVert k\rVert}\Big|^{p}\]
In February 2020, \cite{rohrer_2020} used a one-dimensional input vector to visually demonstrate that an SCS kernel was a more accurate feature detector than a standard convolution kernel. Specifically, Rohrer showed that the SCS kernel more consistently output large values in the presence of a desired feature and smaller values in the absence of such a feature, than a standard convolution kernel. 
The math and examples provided by \cite{rohrer_2020} are convincing that standard convolutions are better filters than detectors, and the hypothesis that such a change from convolutions to SCS would improve predictive accuracy is intuitive and logical. 

In this study, we more thoroughly investigate the SCS, exploring accuracy, efficiency, interpretability, and adversarial robustness as hypotheses of what may be improved by SCS. Our experiments are repeated using multiple different current network architectures to see if any benefits of SCS improve with smaller network size, as hypothesized by current small-scale results of Rohrer. Our results find some measurable differences between SCS and convolutional networks, but said differences do not yet rise to a level that would meaningfully impact current practitioners. 




\section{Review of Related Work}
Since the \cite{rohrer_2020} Twitter thread, many have experimented with various implementations of the SCS layer. Initial results suggest that SCS is very parameter-efficient, which may be desirable in use-cases with limited compute or wattage capacities. On MNIST, \cite{pisoni2022sharpenedcossim} reported achieving 99\% accuracy using an SCS-based architecture with less than 1.4K parameters. \cite{pisoni2022sharpenedcossim} also observed that SCS networks tended to produce more interpretable weights, work well with unscaled inputs, and require no normalization.  It was also observed that SCS kernels could extract not only exact matches of desired features, but also their opposites (i.e., with opposite sign). As such, \cite{pisoni2022sharpenedcossim} recommended that SCS be paired with MaxAbsPool2d (taking the maximum of the absolute values of an input), in contrast to the traditional MaxPool2d, for optimal performance. On CIFAR-10, Rohrer published an SCS-based model in his \texttt{scs-gallery} (see \cite{brohrer_2022}) that claimed to achieve 80\% accuracy with only 25.2K parameters, an order of magnitude less than existing architectures with comparable accuracy. \cite{nestler_2022} experimented with scaling up SCS networks to larger datasets and leveraging TPU computations, but found that SCS was computationally significantly slower to train than convolutional layers. This is primarily because the exponentiation to the $p^{th}$ power in SCS does not parallelize well to GPUs and TPUs.

 In addition to the properties discussed above, members of the machine learning community have found that the inclusion of SCS layers does not appear to produce more accurate models than existing architectures (see \cite{rohrer_2022github}). Others have also found that SCS architectures do not require nonlinear activations (e.g., ReLU, sigmoid), dropout layers, nor normalization layers (e.g., BatchNorm2d) after SCS layers. Some researchers have also tried incorporating SCS into existing architectures such as ResNet (see \cite{batchelor_2022}), compact transformers (see \cite{walton_2022}), models for ASL classification (see \cite{wagner_2022}), and GANs (see \cite{zimonitrome_2022}), with limited or mixed reported results. However, to the best of our knowledge, there do not exist any formally published manuscripts on SCS experiments and behavior, especially in comparison to traditional convolutional layers.

\section{Materials and Methods}

\subsection{Architectures}
For initial exploratory experiments, we selected two network architectures to serve as our test platforms: first, an SCS network with 100K parameters from Rohrer's \texttt{scs-gallery}, which we will call ``Rohrer100K" (see \cite{brohrer_2022}); second, the ResNet18 network from PyTorch's \texttt{torchvision} library with 11.6 million parameters. A ResNet18/20 architecture specially designed for CIFAR-10 was also briefly used when comparing the weight behaviors of SCS layers versus their convolutional counterparts (see \cite{Idelbayev18a}). For follow-on experiments, we selected four more originally SCS-based architectures from Rohrer's \texttt{scs-gallery} that we will call Rohrer25K, Rohrer47K, Rohrer68K, and Rohrer583K (named after their parameter counts, see \cite{brohrer_2022}). The first two of these were reported to achieve 80\%+ accuracy on CIFAR-10, while the last two were reported to achieve 90\%+ accuracy on standard CIFAR-10. These models were selected to explore SCS behavior across a wide range of parameter-counts. For the above architectures, our benchmark was the standard $32 \times 32$ CIFAR-10 dataset. To explore the behavior of SCS on higher-resolution images, we also explored SCS-based ResNet18 variants on CIFAR-10 resized to $224 \times 224$.

For each starting architecture, we tested every variant combination of the following settings: convolutional layer vs. SCS; MaxPool2d vs. MaxAbsPool2d; ReLU vs. no activation. For our follow-up experiments on ResNet18 ($224 \times 224$), in addition to the settings specified above, we also explored variants with or without BatchNorm2d. When replacing one feature with another (e.g., convolutional with SCS), every attempt was made to carry over as many settings as possible from the original starting architecture (e.g., kernel size, number of filters, etc.). In addition, for each architecture family, all model variants using that architecture were trained with the same initial weights. We hope that such practices allow us to more accurately assess the direct effects of transplanting SCS into existing architectures. Please see \autoref{sec:add_exp_details} for additional experimental details.

For all model variants, train and test accuracies, losses, and compute times (in seconds) were recorded per epoch. The norms of the weights and gradients in each layer were also recorded per epoch for all variants. Vanilla gradient saliency maps were generated for each model variant to probe model interpretability. In addition, projected gradient descent (PGD) adversarial attacks were performed against all model variants to further assess model interpretability. For variants using SCS layers, we also recorded the $p$ exponentiation parameter values of each SCS layer per epoch.

\section{Initial Exploratory Experiments on Rohrer100K and ResNet18}
\label{sec:init_explore}
We begin by exploring the accuracy and training efficiency, the nature and parameter behavior, and the interpretability and adversarial robustness of SCS when used in the Rohrer100K and ResNet18 architectures on $32 \times 32$ CIFAR-10. Initial testing suggested that SCS networks tended to require more epochs of training to achieve their maximum test accuracy. As such, all models using architectures originally from \texttt{scs-gallery} were trained for 800 epochs, while all other models were trained for 200 epochs.
Please see \autoref{subsec:init_exp_details} for more experimental setting details. Additional initial exploratory experiments regarding SCS parameter behavior can be found in \autoref{sec:add_init_experiments}.

\subsection{Accuracy and Efficiency}

\begin{table}[h]
    \caption{Test accuracies of initial Rohrer100K model variants on $32 \times 32$ CIFAR-10. See Appendix B for ResNet18 results.}
    \centering
    \begin{adjustbox}{width=1\textwidth}
    \begin{tabular}{llllllllll}
    \hline
        \textbf{Architecture} & \textbf{Layer} & \textbf{Activation} & \textbf{Pooling} & \textbf{Dropout} & \textbf{Normalization} & \textbf{Image Dim.} & \textbf{Val. Acc.} & \textbf{Train Time (s)} & \textbf{Eval Time (s)} \\ \hline
        Rohrer100K & Conv2d & None & MaxAbsPool2d & None & None & 32x32 & 0.4952 & 22.176 & 2.113 \\ 
        Rohrer100K & Conv2d & None & MaxPool2d & None & None & 32x32 & 0.7301 & 21.997 & 2.12 \\
        Rohrer100K & Conv2d & ReLU & MaxAbsPool2d & None & None & 32x32 & 0.6496 & 22.134 & 2.118 \\ 
        Rohrer100K & Conv2d & ReLU & MaxPool2d & None & None & 32x32 & 0.8032 & 22.118 & 2.126 \\
        Rohrer100K & SCS & None & MaxAbsPool2d & None & None & 32x32 & 0.7761 & 33.482 & 1.871 \\ 
        Rohrer100K & SCS & None & MaxPool2d & None & None & 32x32 & 0.787 & 33.341 & 1.791 \\ 
        Rohrer100K & SCS & ReLU & MaxAbsPool2d & None & None & 32x32 & 0.777 & 33.523 & 1.817 \\
        Rohrer100K & SCS & ReLU & MaxPool2d & None & None & 32x32 & 0.8026 & 33.381 & 1.776 \\ 
        \hline
    \end{tabular}
    \end{adjustbox}
    \label{table:rohrer100k_initial_accuracies}
\end{table}

From Table \ref{table:rohrer100k_initial_accuracies} above and Table \ref{table:ResNet18_initial_accuracies} (see \autoref{sec:init_exp_tables}), we see that within the same architecture family, the use of SCS layers does not yield a significant increase in accuracy (approx. $78\%$ for Rohrer100K and $82\%$ for ResNet18) on CIFAR-10. However, we do confirm that SCS is able to achieve comparable accuracy performance to standard convolutional layers without the inclusion of ReLU nonlinear activation functions, while convolutional layers must be paired with such nonlinear activations for maximum performance. Yet, it appears that the inclusion of ReLU activations in SCS networks may still provide a small increase in accuracy performance. Interestingly, in Rohrer100K, we also find that standard convolutional layers paired with MaxAbsPool2d (and ReLU) incur a significant dip in accuracy performance. But, this dip does not seem to occur with ResNet18.

We found that SCS-based networks require significantly more time (seconds per epoch) to train than their corresponding convolution-based counterparts. This corroborates best practices and observations that others have found. However, while all SCS-based ResNet18 variants were slower in evaluation time than their convolution-based counterparts, all SCS-based Rohrer100K variants were noticeably faster in evaluation time than their convolution-based counterparts. This suggests that SCS properties may not unilaterally hold across all architectures.
\subsection{Cosine Similarity (Unsharpened)}
We hypothesize that SCS does not require nonlinear activations like ReLU to achieve peak performance because the exponentiation by $p$ (sharpening) in SCS is effectively an activation function in and of itself. In other words, we postulate that SCS could be interpreted as a combination of output normalization (via cosine similarity) and an exponential activation function. To test this hypothesis, we fix $p=1$ in the SCS layers of Rohrer100K to test whether (unsharpened) cosine similarity on its own would require an activation function. The maximum accuracy of cosine similarity coupled with ReLU was $76.04\%$, while the maximum accuracy of cosine similarity without ReLU was $56.58\%$. 

\subsection{Weight Norms of SCS Layers}
To explore how the normalization and exponentiation components of SCS would affect the weights used in its convolution component (versus a standard convolutional layer), we compare the weights of a fully-trained convolution-based network versus those of a fully-trained SCS-based network. To ensure that the weights of our fully-trained convolution-based network represented the maximum performance (reasonably) possible, we decided to use a ResNet18 architecture specifically-optimized for CIFAR-10 by Idelbayev (see \cite{Idelbayev18a}), as opposed to the \texttt{torchvision} variant optimized for ImageNet. We trained two SCS-based variants, each only replacing the convolutional layers for SCS, and not altering any other model components. However, one variant was trained using randomly-initialized starting weights, while the other was trained using pre-trained weights specifically designed for convolution-layers on CIFAR-10. We recorded the L2 norms of each SCS layer's convolution weights over each mini-batch.

From Figure \ref{fig:idelbayev_weight_norms_comparison} in \autoref{sec:init_exp_tables}, we see that for the variant trained from scratch, the L2 norms of each SCS layer's weights did not appear to noticeably change over epoch time. However, for the variant trained from pre-trained weights originally optimized for standard convolution layers, the L2 norms of the SCS layer weights significantly monotonically increased over mini-batches. As such, these results suggest that SCS-based architectures prefer larger weights than their convolution-based counterparts. The L2 norms reported at mini-batch 0 in the pre-trained plot are precisely the norms found in the fully-trained convolution-based network. It is possible that SCS may tolerate larger weights because of the output normalization. Perhaps larger weights may also contribute to more sensitive feature detection and differentiation. 

\subsection{Saliency Maps}
To investigate Pisoni's observations regarding SCS interpretability, vanilla gradient saliency maps were generated for convolution and SCS variants of Rohrer100K and ResNet18, presented as Figures \ref{fig:rohrer100k_initial_saliency} and \ref{fig:ResNet18_initial_saliency} in \autoref{sec:init_exp_tables}. In Figure \ref{fig:rohrer100k_initial_saliency_BEST} below, we provide one representative example of the saliency maps of a convolution-based and an SCS-based Rohrer100K corresponding to an airplane (class 0). We observe that the convolution-based variant yields a ``scattered" saliency map, while the SCS variant yields a more focused map that corresponds to the triangular-shaped aircraft. Overall, across Rohrer100K and ResNet18, our initial results suggest that SCS-based model variants tend to learn representations that are more interpretable and focus on more critical parts of the image than their convolution-based counterparts.
\begin{figure}[t]
    \centering
    \subfloat[\centering With convolution]{{\includegraphics[width=2cm]{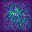} }}%
    \qquad
    \subfloat[\centering With SCS]{{\includegraphics[width=2cm]{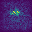} }}%
    \qquad
    \subfloat[\centering Original image]{{\includegraphics[width=2cm]{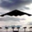} }}%
    \caption{Saliency maps of best convolutional and SCS initial Rohrer100K variants on CIFAR-10. The left panel shows the convolutional variant with the highest class-0-specific accuracy ($82\%$): convolutions, with ReLU and MaxPool2d. The right panel shows the SCS variant with the highest class-0-specific accuracy ($84\%$): SCS, also with ReLU and MaxPool2d.}%
    \label{fig:rohrer100k_initial_saliency_BEST}%
\end{figure}
\subsection{Robustness to Adversarial Attacks}
To further explore the interpretability of SCS-based networks, we performed projected gradient descent (PGD) adversarial attacks on all Rohrer100K and ResNet18 model variants, varying attack strength from $\epsilon=0.001$ to $\epsilon=0.030$. We posit that architectures with more interpretable representations (i.e., focus more on true features and signals, and less noise) should be more robust to adversarial attacks. From Figures \ref{fig:rohrer100k_initial_saliency} and \ref{fig:ResNet18_initial_saliency} in \autoref{sec:init_exp_tables}, we find that SCS-based Rohrer100K and ResNet18 variants are noticeably more adversarially robust than their convolution-based counterparts, in terms of experiencing a slower decay in accuracy with increasing attack strength. The one outlier in the Rohrer100K trials was the convolution-based variant with MaxAbsPool2d and ReLU, which was also an outlier in terms of achieving only 64.96\% accuracy (while other variants achieved high 70\%s or 80\%s). Notably, for ResNet18, the best SCS-based ResNet18 variant (incidentally, using the best practices suggested by Rohrer and the ML community) was particularly more adversarially robust than all convolution-based variants. As such, the initial adversarial experimental evidence further suggests that the incorporation of SCS layers may produce more interpretable networks.

\section{Follow-up Experiments}
To investigate the generalizability of our initial SCS findings across different-sized architectures, we repeated our previous experiments on four additional models from the \texttt{scs-gallery}: Rohrer25K, Rohrer47K, Rohrer68K, Rohrer100K (with new randomized initial weights), and Rohrer583K. To further discern which properties of SCS can be attributed to normalization and which to exponentiation, we include a third potential layer option: an ablation of convolutions and SCS that we will call SharpenedSDP. SharpenedSDP stands for "sharpened strided dot product," and is mathematically equivalent to a standard convolutional layer raised to a learned $p$ exponent, \textit{without} the normalization. From our initial testing, we recalled that using convolutional layers without ReLU would almost always lead to poor performance. As such, we no longer tested this combination in the following experiments. To investigate SCS on higher-resolution images, we also repeated these experiments on ResNet18 with CIFAR-10 resized to $224 \times 224$. In addition to the settings described above, we included variants with/without BatchNorm2d (present in the original Resnet18 architecture). We also tried SharpenedSDP with $224 \times 224$ ResNet18, but all of these variants encountered vanishing and/or exploding gradients, so we omit them from discussion.


\subsection{Accuracy and Efficiency}
From Tables \ref{table:rohrer25k_accuracies}-\ref{table:rohrer583k_accuracies} in \autoref{sec:followup_exp_tables}, we find that our initial conclusions regarding SCS accuracy and efficiency generally hold strong across Rohrer25K through Rohrer583K. We also observe that the SharpenedSDP variants do not require activation functions, further corroborating our hypothesis that the exponentiation to the $p^{th}$ power is a substitute for nonlinear activation functions. In addition, we found that SCS and SharpenedSDP variants appear to consistently perform faster than their convolution-based counterparts on evaluation time. This is potentially one major advantage of SCS: while SCS-based networks might be slower to train, they are faster during evaluation. Finally, across Rohrer 25K through Rohrer583K, we found that model variants containing both ReLU and MaxAbsPool2d will almost always perform poorly, regardless of whether we are using SCS, SharpenedSDP, or standard convolutional layers. From Table \ref{table:ResNet18_224_accuracies} in \autoref{sec:followup_exp_tables},  we find that our initial conclusions regarding SCS accuracy and efficiency also generally hold strong on $224 \times 224$ ResNet18. However, in contrast to Rohrer25K through Rohrer583K, we do not see a dip in performance associated with the inclusion of both MaxAbsPool2d and ReLU. We also observed that the inclusion of BatchNorm2d, holding everything else constant, appears to grant a 3-4\% increase in accuracy.

\subsection{Saliency Maps and Interpretability}
From the saliency maps for Rohrer25K through 583K, as shown in Figures \ref{fig:rohrer25k_saliency}-\ref{fig:rohrer583k_saliency} in \autoref{sec:followup_exp_tables}, we see that our initial findings regarding SCS-based model variants learning sparser and more interpretable representations appear to hold strong. In addition, we find significant evidence corroborating the hypothesis that the $p$ exponentiation component in SCS is what creates such ``sparse" saliency maps. We conclude as such because the SharpenedSDP saliency maps much more consistently resembled the SCS saliency maps than the convolution ones. From Figure \ref{fig:ResNet18_224_saliency} in \autoref{sec:followup_exp_tables}, we find that our conclusions regarding SCS versus convolutional variants' saliency maps also extend to higher-resolution $224 \times 224$ images.

\subsubsection{Adversarial Robustness}
Repeating the PGD attack simulations on Rohrer25K - 583K, we found our results differing significantly from our initial Rohrer100K results. This was perplexing, because the underlying constructions of all the \texttt{scs-gallery} models were all very similar. From Figure \ref{fig:rohrer_family_PGD} in \autoref{sec:followup_exp_tables}, we find that the SharpenedSDP/MaxAbsPool2d/ReLU and convolution/MaxAbsPool2d/ReLU variants consistently outperformed the top SCS-based variant.  However, these two variants also consistently performed significantly lower in terms of accuracy, across all four \texttt{scs-gallery} architectures (sometimes only hitting 10\% accuracy).
To verify our observations, we simulated PGD attacks against Rohrer100K again, this time using a different set of randomized initial weights (see Figure \ref{fig:rohrer100k_pgd_LATE} in \autoref{sec:followup_exp_tables}). Again, we found that SharpenedSDP/MaxAbsPool2d/ReLU and convolution/MaxAbsPool2d/ReLU were the top adversarially-robust variants (though their original accuracies were dismal). There did not seem to be a significant difference between the remaining SCS, SharpenedSDP, and convolution-based variants, but the top performers with decent accuracy appear to be SharpenedSDP/MaxAbsPool2d/noReLU, followed by SCS/MaxAbsPool2d/ReLU.

PGD attacks on $224 \times 224$ ResNet18 variants also present a different narrative than the initial PGD attacks on the $32 \times 32$ ResNet18 variants. As shown in Figure \ref{fig:pgd_ResNet18_224} in \autoref{sec:followup_exp_tables}, for variants without BatchNorm2d, convolution-based variants were noticeably more robust to adversarial attack. However, with BatchNorm2d, the best SCS-based variant maintains a small but consistent advantage over the convolution-based variants. The differences are not nearly as pronounced as in the $32 \times 32$ CIFAR-10 experiments, however. Upon closer inspection, it appears that the inclusion or exclusion of BatchNorm2d simply does not affect SCS-based variants' adversarial robustness, but does significantly affect convolution-based variants' robustness. The causes of this behavior are not immediately obvious, and further exploration is necessary.
\section{Discussion and Future Work}
 In this manuscript, we find strong evidence supporting the following five generalizations about SCS performance and behavior. First, SCS does not yield significantly higher accuracy than standard convolution based layers. Second, while SCS-based models are almost always slower to train (seconds per epoch) than their convolution-based counterparts, they frequently benefit from slightly faster evaluation times (seconds per epoch). Third, SCS can be interpreted as a combination of convolution, normalization, and nonlinear exponential activation. As such, they are capable of performing excellently without additional normalization or nonlinear activations like ReLU. Fourth, SCS tends to yield more interpretable learned representations than convolutions. Fifth, in certain environments, SCS also yields potentially more adversarially robust networks.

 However, the above properties were not absolute. This suggests that characterizing SCS performance may be more analytically and computationally-involved than solely swapping out convolutional layers and associated components for SCS layers. It appears that SCS performance and properties can vary noticeably, depending on the existing architecture and other components present in the system. Significant future work will be required to establish accompanying best practices for SCS networks and potentially construct novel architectures that best leverage SCS's unique properties. For example, creating a ResNet bottleneck module optimized for SCS may not be as straightforward as swapping out convolutional layers for SCS and replacing MaxPool2d with MaxAbsPool2d.

 Given (unsharpened) cosine similarity's established uses in text modeling, a natural applied extension is to transplant SCS into existing text modeling architectures. SCS's interpretability properties also warrant its transplanting to malware classification models that currently use standard convolutions. Another direction is to explore the use of SCS in convolution-based transformer models. We conclude that SCS is a promising alternative to convolutional layers, with its own set of unique advantages and drawbacks.

\bibliography{references,RaffReferences}

\newpage

\appendix

\section{Additional Experimental Details}
\label{sec:add_exp_details}

To standardize model training, unless otherwise specified, we used an Adam optimizer, coupled with a maximum learning rate of $0.01$ and a \texttt{OneCycleLR} learning rate scheduler (see \cite{smith_2017}). RandomCrop and RandomHorizontalFlip data augmentations were also applied to the training set. Such settings were directly borrowed from training scripts in Rohrer's \texttt{scs-gallery}. Initial experiments on Rohrer100K suggested that weight decay did not noticeably affect SCS-based architectures' performance, so for all results presented, weight decay was set to $0$. For reproducibility~\cite{Raff2022a,Raff2020c,Raff2019_quantify_repro} and to rule out the effects of different weight initializations on model performance, we ensured that all variants of the same base architecture (e.g., ResNet18 on $32 \times 32$) were trained with the same set of randomly-initialized weights. 

All experiments were performed using PyTorch 1.11/1.12 on Linux. The hardware available were 8 NVIDIA Tesla P100 GPUs with 16 GB of RAM each, and 8 NVIDIA Tesla V100 GPUs with 32 GB of RAM each. We gratefully acknowledge the use of the Sharpened Cosine Similarity and MaxAbsPool2d PyTorch implementations borrowed from Rohrer's public repository (see \cite{rohrer_2022github}).

\subsection{Additional Details on Initial Exploratory Experiments}
\label{subsec:init_exp_details}

We only tested combinations of convolution or SCS, ReLU or no activation, and MaxPool2d or MaxAbsPool2d. We retained the BatchNorm2d and AdaptiveAvgPool layers in all tested ResNet18 variants in this section.

The accuracy results for ResNet18 in \autoref{sec:init_explore} were obtained using a \texttt{torchvision} ResNet18 originally designed for ImageNet, but we replaced the final fully-connected layer to account for the 10 classes in CIFAR-10. During initial experimentation, we also tried transfer learning with SCS networks by starting ResNet18 training using pre-trained ResNet18 weights for ImageNet. We found that there was no significant difference between model variants trained from scratch or using pre-trained weights. As such, we only report results using randomly-initialized starting weights.

\section{Tables and Additional Figures from Initial Exploratory Experiments}
\label{sec:init_exp_tables}


\begin{table}[h]
    \caption{Test accuracies of initial ResNet18 model variants on $32 \times 32$ CIFAR-10.\newline}
    \label{table:ResNet18_initial_accuracies}
    \centering
    \begin{adjustbox}{width=1\textwidth}
    \begin{tabular}{llllllllll}
    \hline
        \textbf{Architecture} & \textbf{Layer} & \textbf{Activation} & \textbf{Pooling} & \textbf{Dropout} & \textbf{Normalization} & \textbf{Image Dim.} & \textbf{Val. Acc.} & \textbf{Train Time (s)} & \textbf{Eval Time (s)} \\ \hline
        ResNet18 & Conv2d & None & MaxAbsPool2d & None & BatchNorm2d & 32x32 & 0.6027 & 28.816 & 2.217 \\
        ResNet18 & Conv2d & None & MaxPool2d & None & BatchNorm2d & 32x32 & 0.7146 & 26.872 & 2.113 \\
        ResNet18 & Conv2d & ReLU & MaxAbsPool2d & None & BatchNorm2d & 32x32 & 0.8225 & 28.067 & 2.245 \\
        ResNet18 & Conv2d & ReLU & MaxPool2d & None & BatchNorm2d & 32x32 & 0.8219 & 26.478 & 2.141 \\
        ResNet18 & SCS & None & MaxAbsPool2d & None & BatchNorm2d & 32x32 & 0.8134 & 86.227 & 3.176 \\
        ResNet18 & SCS & None & MaxPool2d & None & BatchNorm2d & 32x32 & 0.8258 & 83.794 & 3.089 \\
        ResNet18 & SCS & ReLU & MaxAbsPool2d & None & BatchNorm2d & 32x32 & 0.8306 & 85.369 & 3.075 \\
        ResNet18 & SCS & ReLU & MaxPool2d & None & BatchNorm2d & 32x32 & 0.8241 & 83.711 & 2.908 \\ \hline
    \end{tabular}
    \end{adjustbox}
\end{table}

\subsection{Weight Norms of SCS Layers}
\begin{figure}[H]
    \centering
    \subfloat[\centering From scratch]{{\includegraphics[width=13.5cm]{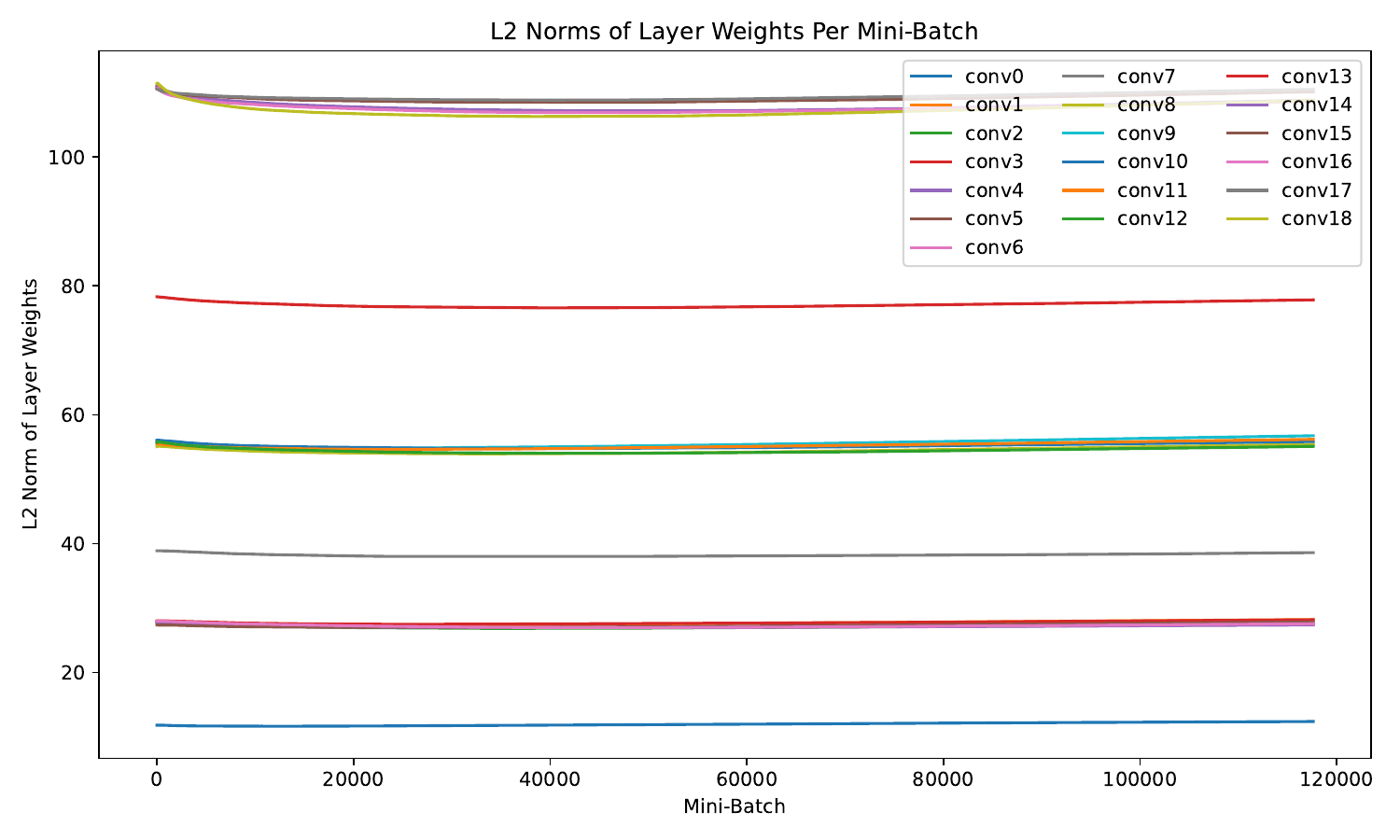} }}%
    \qquad
    \subfloat[\centering From pre-trained weights for \texttt{nn.Conv2d}]{{\includegraphics[width=13.5cm]{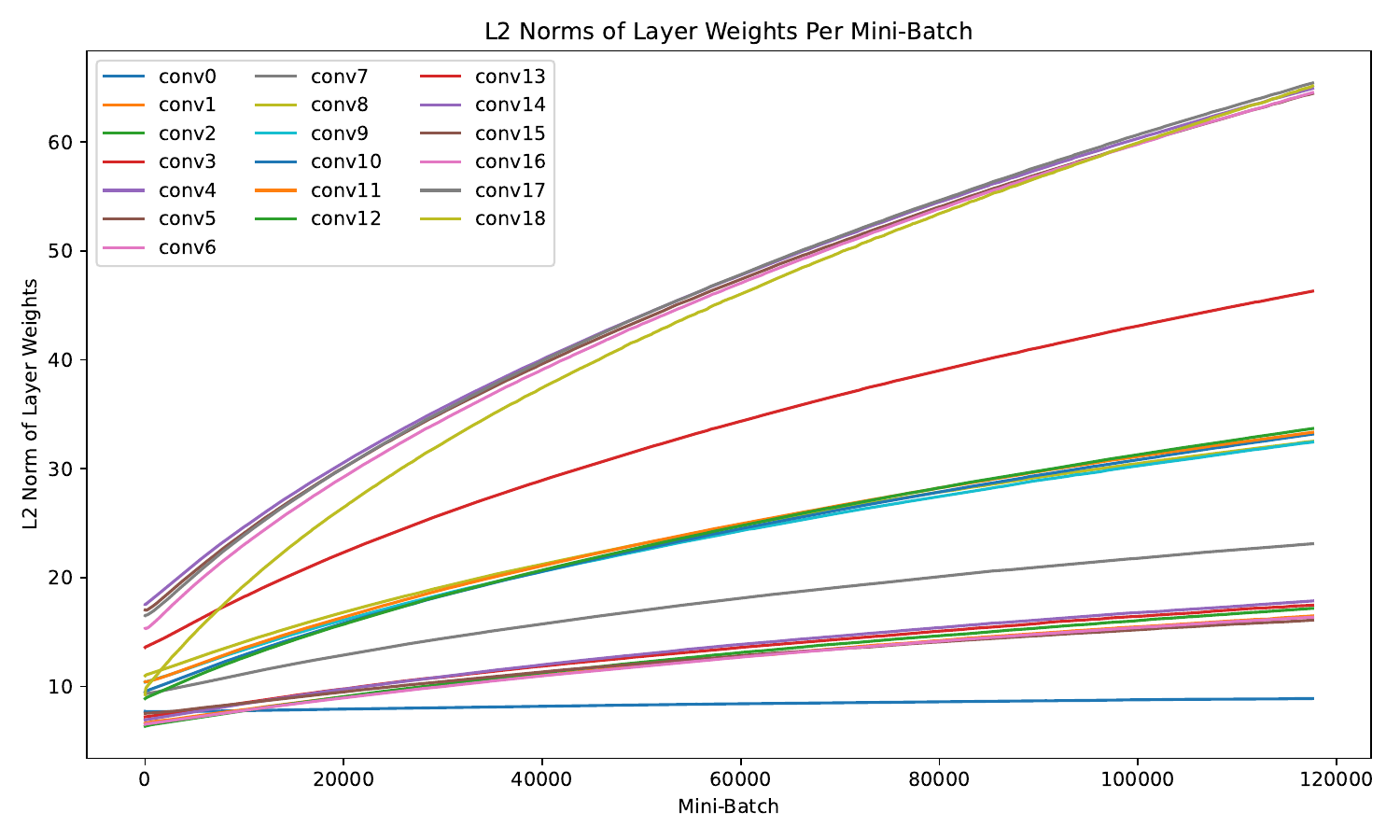} }}%
    \caption{Magnitudes of Idelbayev ResNet18 weights over mini-batches.}%
    \label{fig:idelbayev_weight_norms_comparison}%
\end{figure}

\subsection{Saliency Maps and Interpretability}
\begin{figure}[H]
\centering
  \subfloat[Saliency Maps]{
	\begin{minipage}[c][0.5\width]{
	   0.75\textwidth}
	   \centering
\includegraphics[width=1\textwidth]{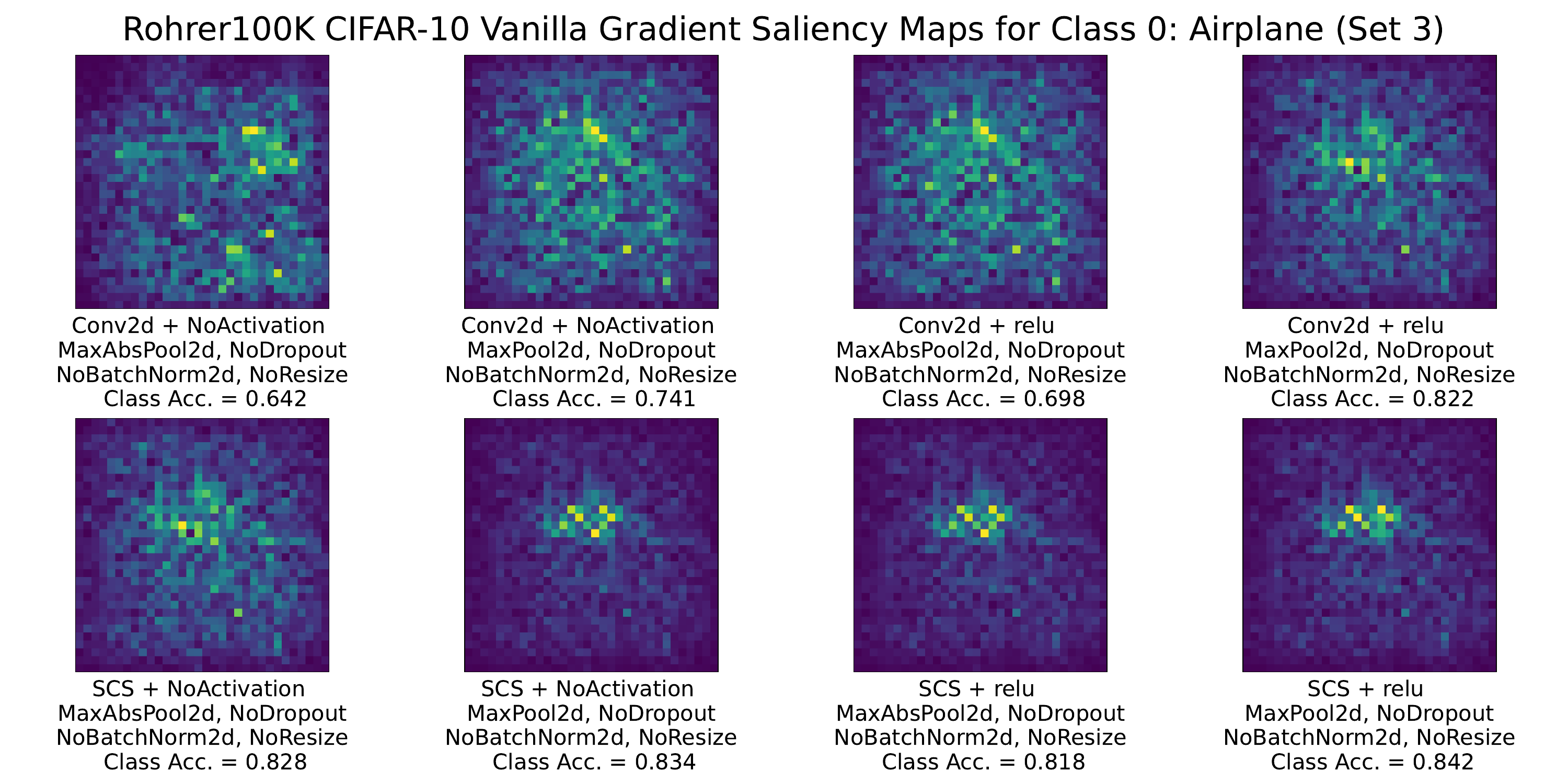}
	\end{minipage}}
  \subfloat[Original Image]{
	\begin{minipage}[c][1.2\width]{
	   0.18\textwidth}
	   \centering
	   \includegraphics[width=1\textwidth]{figures/saliency_maps/saliency_map_rohrer100k_initial_reference.png}
	\end{minipage}}
\caption{Saliency maps of initial Rohrer100K variants on CIFAR-10.}
\label{fig:rohrer100k_initial_saliency}
\end{figure}

\begin{figure}[H]
\centering
  \subfloat[Saliency Maps]{
	\begin{minipage}[c][0.5\width]{
	   0.75\textwidth}
	   \centering
\includegraphics[width=1\textwidth]{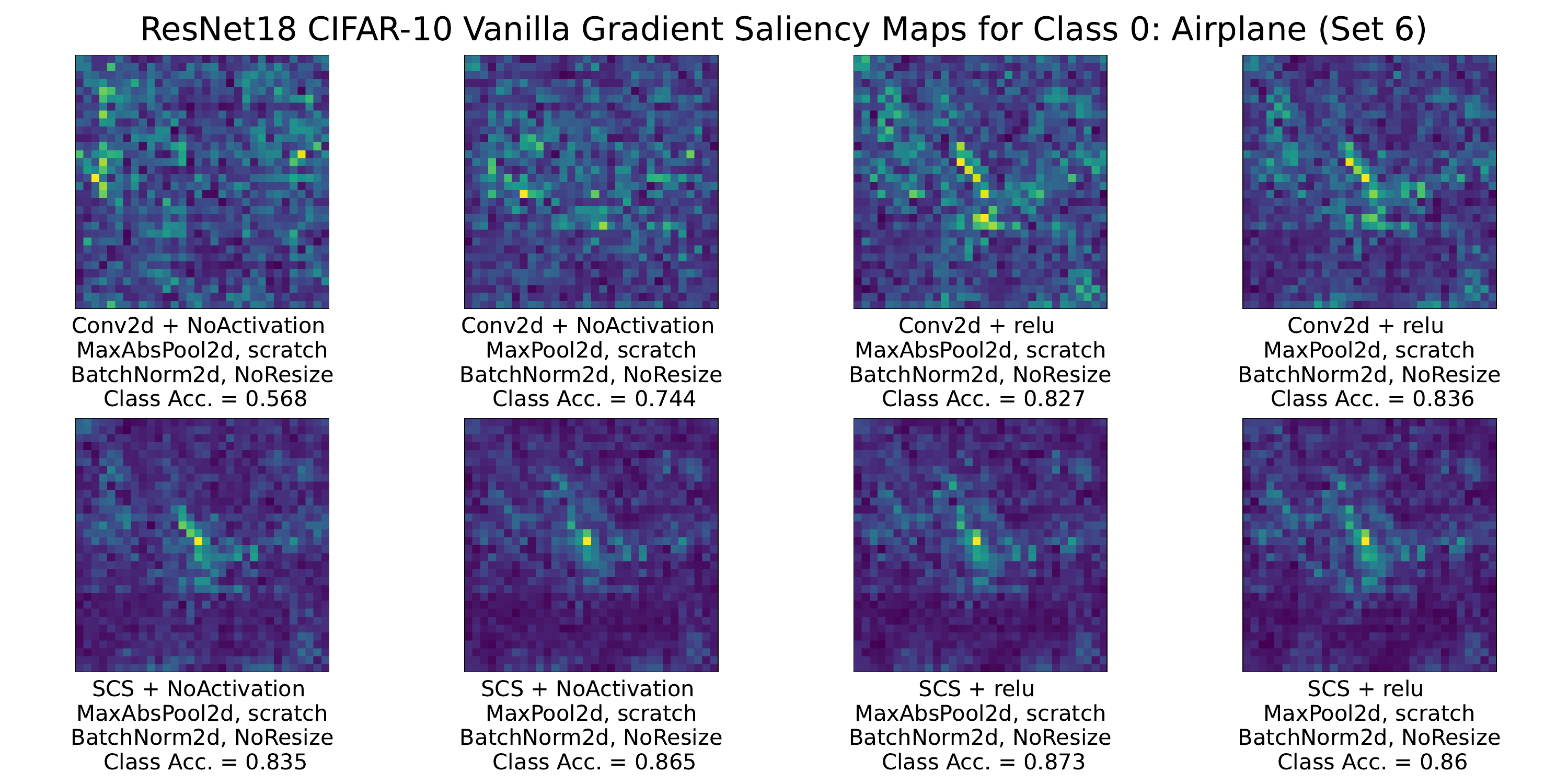}
	\end{minipage}}
  \subfloat[Original Image]{
	\begin{minipage}[c][1.2\width]{
	   0.18\textwidth}
	   \centering
	   \includegraphics[width=1\textwidth]{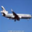}
	\end{minipage}}
\caption{Saliency maps of initial ResNet18 variants on CIFAR-10.}
\label{fig:ResNet18_initial_saliency}
\end{figure}

\subsection{Adversarial Robustness}
\begin{figure}[H]
     \centering
     \includegraphics[scale = 0.35]{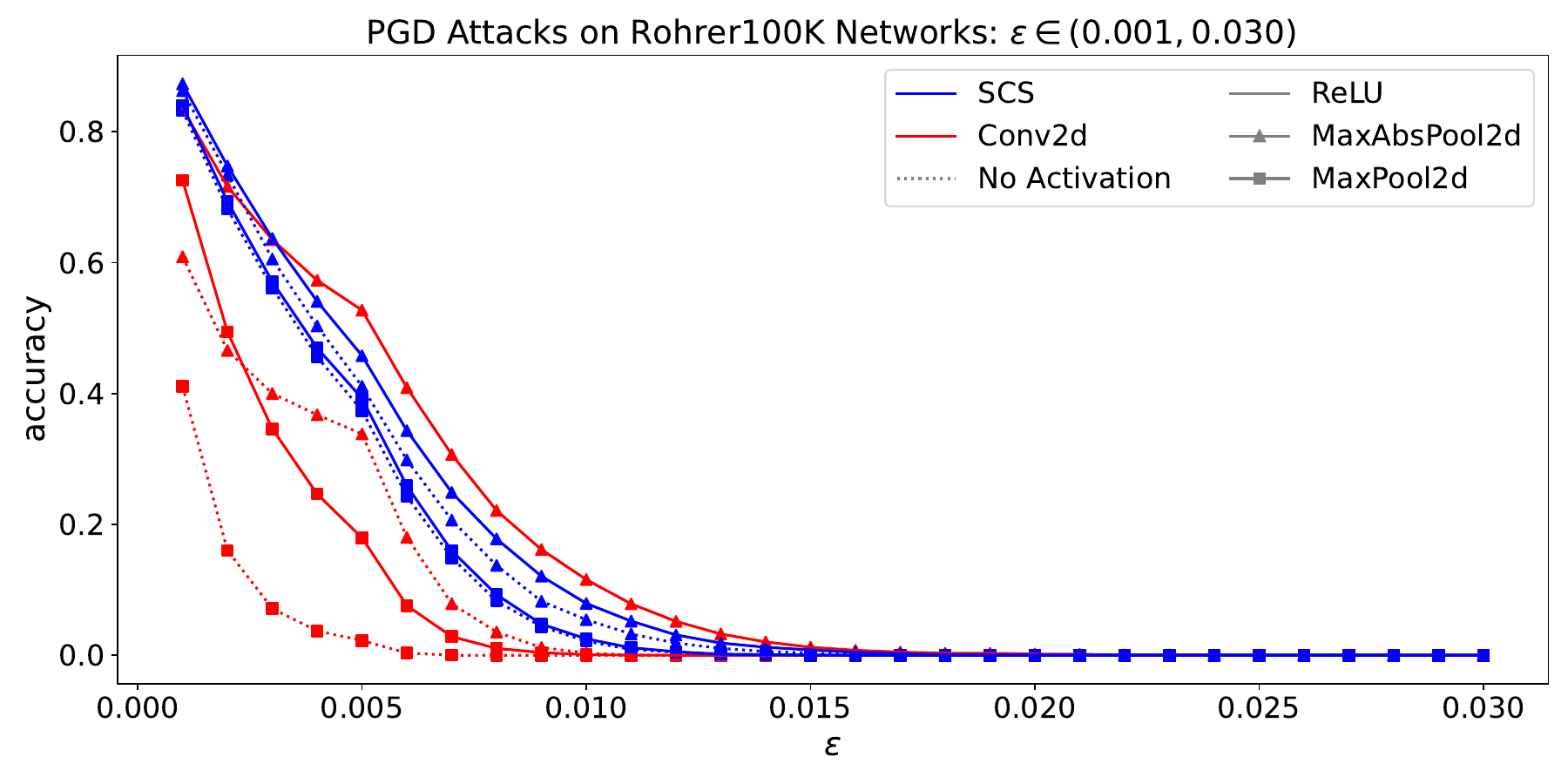}
     \caption{PGD robustness of Rohrer100K variants on CIFAR-10 ($32 \times 32$, initial testing).}
     \label{fig:rohrer100k_pgd_early}
\end{figure}

\begin{figure}[H]
     \centering
     \includegraphics[scale = 0.35]{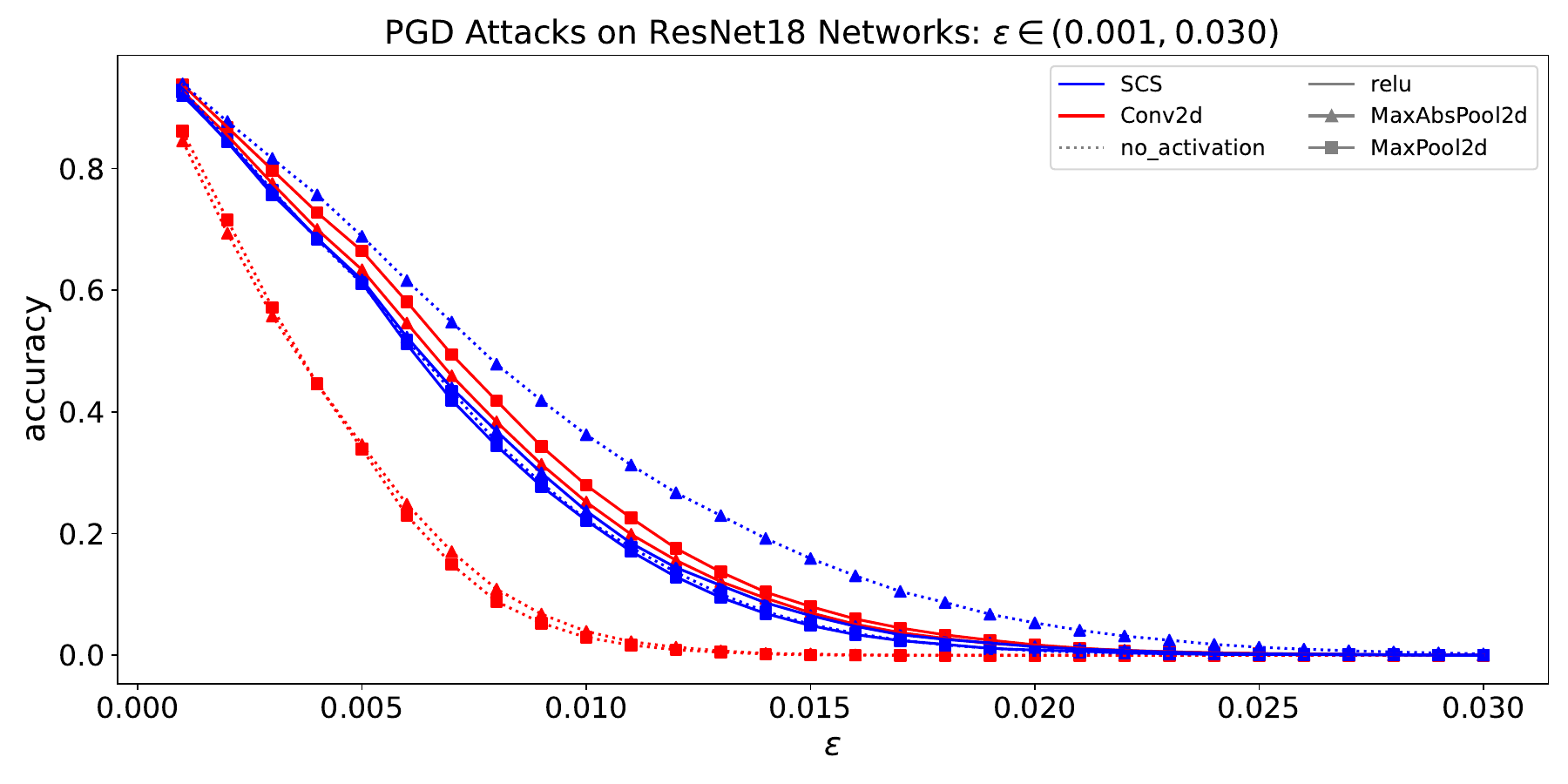}
     \caption{PGD robustness of ResNet18 variants on CIFAR-10 ($32 \times 32$, initial testing).}
     \label{fig:ResNet18_pgd_early}
\end{figure}

\section{Additional Initial Experiments}
\label{sec:add_init_experiments}

\subsection{P-values of SCS Layers}
\label{subsec:scs_pval}
We also explored the behavior of the $p$ values in a SCS layer over epoch time. Using a fixed learning rate of $1.0 \times 10^{-3}$, coupled with a weight decay of $1.0 \times 10^{-5}$, we plotted the learned values of $p$ for the kernels in the first SCS layer of Rohrer100K (with MaxAbsPool2d, without ReLU). From Figure \ref{fig:rohrer100k_pvals} below, we find that there are not many dynamics of note. Monotonicity does not appear to hold.
\begin{figure}[H]
     \centering
     \includegraphics[scale = 0.5]{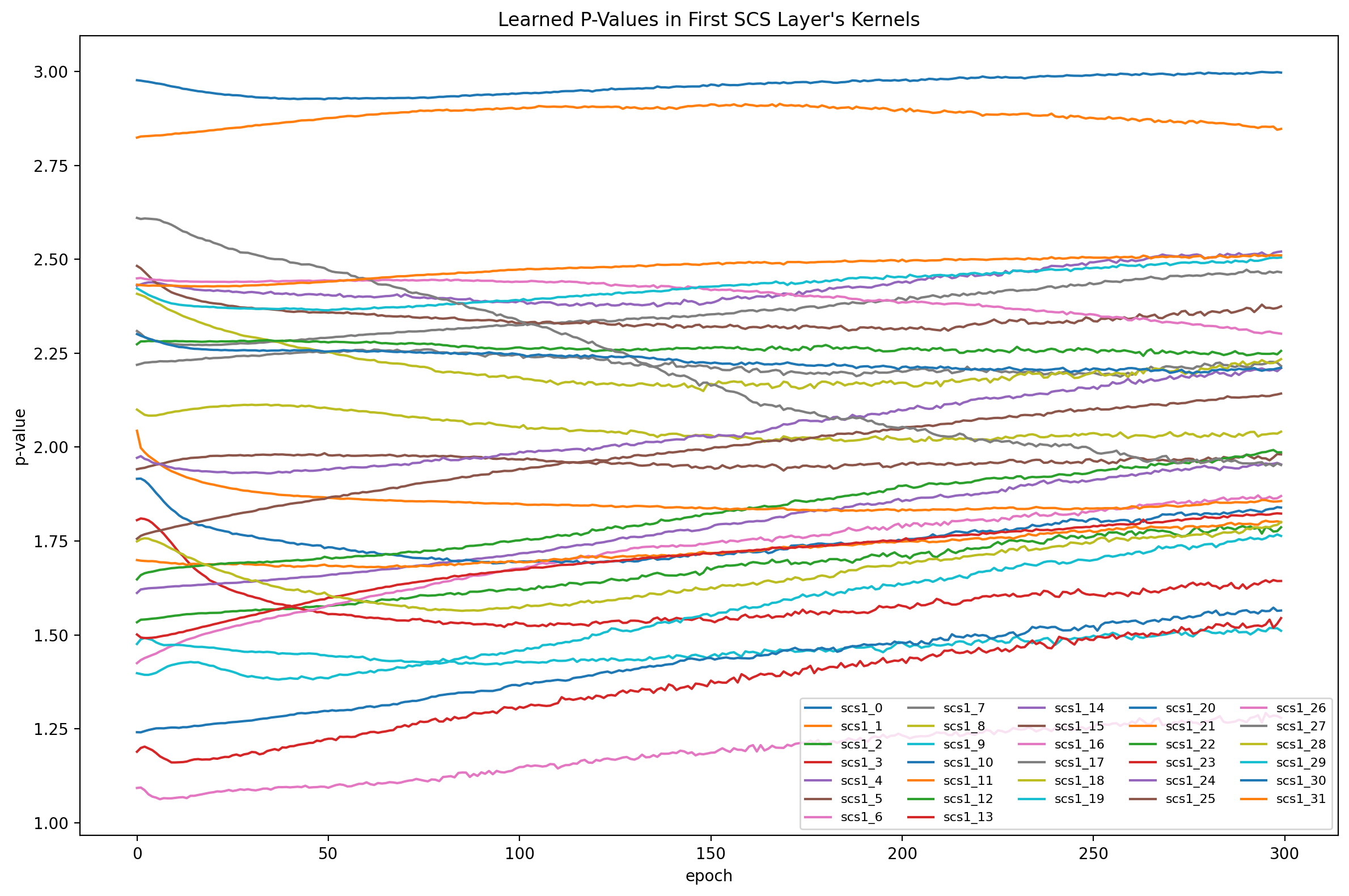}
     \caption{Learned values of $p$ in first SCS layer for Rohrer100K on CIFAR-10 ($32 \times 32$, Initial Testing).}
     \label{fig:rohrer100k_pvals}
\end{figure}

\subsection{Optuna Hyperparameter Tuning of Fixed-\textit{p} Hyperparameter}
We also wished to investigate whether certain values of $p$ were more amenable to SCS performance than others. As a proxy, we used a custom SCS layer with $p$ fixed to a certain specified value. We swapped out the original layers in both Rohrer100K and \texttt{torchvision}'s ResNet18 for these fixed-$p$ SCS layers, and preserved all other existing components. For simplicity, we forced all SCS layers in each network to share the same initialized and fixed value of $p$. We used the \texttt{Optuna} Bayesian hyperparameter tuning utility to find the most optimal values of $p$ (see \cite{optuna_2019}).  Since multiple initial values of $p$ needed to be tested, we restricted each Rohrer100K trial to 600 epochs (versus the standard 800 epochs) and each ResNet18 trial to 100 epochs (versus the standard 200 epochs). We also enabled pruning to terminate non-promising $p$-value trials early.
\begin{figure}[H]
    \centering
    \subfloat[\centering Rohrer100K]{{\includegraphics[width=9.5cm]{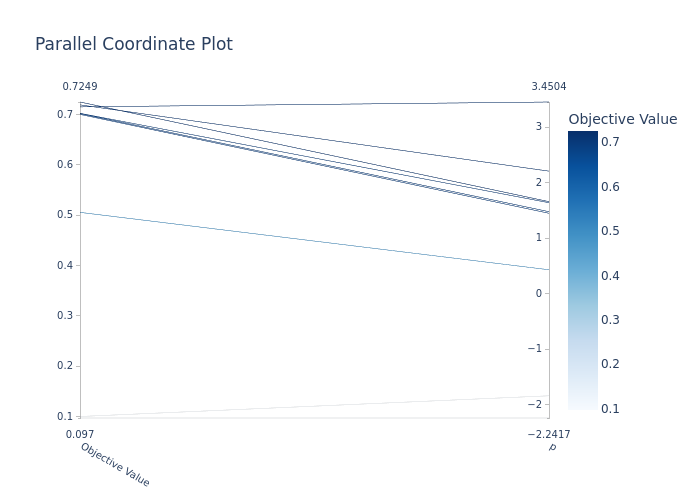} }}%
    \qquad
    \subfloat[\centering ResNet18]{{\includegraphics[width=9.5cm]{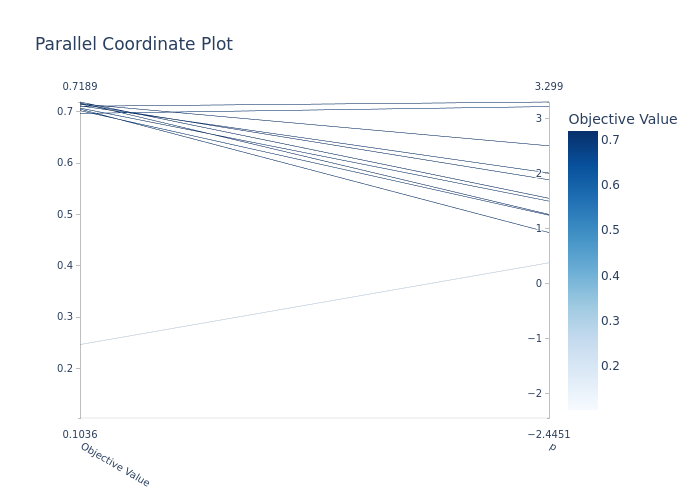} }}%
    \caption{Optuna parallel coordinate plots of $p$ value and accuracy.}%
    \label{fig:optuna}%
\end{figure}

 Optuna results across both Rohrer100K and ResNet18 suggest that all tested values of $p$ above or near $1.0$ appear to yield good performance. This suggests that the specific value of $p$ may not be the most important -- rather, it is the mere presence of the (increasing) exponentiation operation that enables successful SCS performance.
 
\section{Tables and Figures From Follow-up Experiments}
\label{sec:followup_exp_tables}

\subsection{Accuracy and Efficiency}
\begin{table}[H]
    \caption{Test accuracies of Rohrer25K model variants on $32 \times 32$ CIFAR-10.\newline}
    \centering
    \begin{adjustbox}{width=1\textwidth}
    \begin{tabular}{llllllllll}
    \hline
        \textbf{Architecture} & \textbf{Layer} & \textbf{Activation} & \textbf{Pooling} & \textbf{Dropout} & \textbf{Normalization} & \textbf{Image Dim.} & \textbf{Val. Acc.} & \textbf{Train Time (s)} & \textbf{Eval Time (s)} \\ \hline
        Rohrer25K & Conv2d & ReLU & MaxAbsPool2d & None & None & 32x32 & 0.1301 & 94.09 & 9.456 \\
        Rohrer25K & Conv2d & ReLU & MaxPool2d & None & None & 32x32 & 0.7609 & 92.674 & 9.379 \\
        Rohrer25K & SCS & None & MaxAbsPool2d & None & None & 32x32 & 0.7745 & 132.628 & 8.104 \\
        Rohrer25K & SCS & None & MaxPool2d & None & None & 32x32 & 0.7705 & 131.594 & 8.092 \\
        Rohrer25K & SCS & ReLU & MaxAbsPool2d & None & None & 32x32 & 0.2867 & 131.643 & 8.2 \\
        Rohrer25K & SCS & ReLU & MaxPool2d & None & None & 32x32 & 0.7682 & 131.877 & 8.132 \\
        Rohrer25K & SharpenedSDP & None & MaxAbsPool2d & None & None & 32x32 & 0.7719 & 127.797 & 8.038 \\
        Rohrer25K & SharpenedSDP & None & MaxPool2d & None & None & 32x32 & 0.7818 & 126.883 & 7.882 \\
        Rohrer25K & SharpenedSDP & ReLU & MaxAbsPool2d & None & None & 32x32 & 0.1 & 128.575 & 8.056 \\
        Rohrer25K & SharpenedSDP & ReLU & MaxPool2d & None & None & 32x32 & 0.7774 & 127.716 & 8.008 \\ \hline
    \end{tabular}
    \end{adjustbox}
    \label{table:rohrer25k_accuracies}
\end{table}

\begin{table}[H]
    \caption{Test accuracies of Rohrer47K model variants on $32 \times 32$ CIFAR-10.\newline}
    \centering
    \begin{adjustbox}{width=1\textwidth}
    \begin{tabular}{llllllllll}
    \hline
        \textbf{Architecture} & \textbf{Layer} & \textbf{Activation} & \textbf{Pooling} & \textbf{Dropout} & \textbf{Normalization} & \textbf{Image Dim.} & \textbf{Val. Acc.} & \textbf{Train Time (s)} & \textbf{Eval Time (s)} \\ \hline
        Rohrer47K & Conv2d & ReLU & MaxAbsPool2d & None & None & 32x32 & 0.1 & 92.919 & 9.233 \\
        Rohrer47K & Conv2d & ReLU & MaxPool2d & None & None & 32x32 & 0.7978 & 92.469 & 9.274 \\
        Rohrer47K & SCS & None & MaxAbsPool2d & None & None & 32x32 & 0.7671 & 123.82 & 8.564 \\
        Rohrer47K & SCS & None & MaxPool2d & None & None & 32x32 & 0.7893 & 123.125 & 8.425 \\
        Rohrer47K & SCS & ReLU & MaxAbsPool2d & None & None & 32x32 & 0.2513 & 125.136 & 8.643 \\
        Rohrer47K & SCS & ReLU & MaxPool2d & None & None & 32x32 & 0.7816 & 124.303 & 8.554 \\
        Rohrer47K & SharpenedSDP & None & MaxAbsPool2d & None & None & 32x32 & 0.7866 & 121.93 & 8.155 \\
        Rohrer47K & SharpenedSDP & None & MaxPool2d & None & None & 32x32 & 0.7919 & 121.683 & 8.017 \\
        Rohrer47K & SharpenedSDP & ReLU & MaxAbsPool2d & None & None & 32x32 & 0.652 & 123.186 & 8.252 \\
        Rohrer47K & SharpenedSDP & ReLU & MaxPool2d & None & None & 32x32 & 0.7972 & 121.878 & 8.154 \\ \hline
    \end{tabular}
    \end{adjustbox}
    \label{table:rohrer47k_accuracies}
\end{table}

\begin{table}[H]
    \caption{Test accuracies of Rohrer68K model variants on $32 \times 32$ CIFAR-10.\newline}
    \centering
    \begin{adjustbox}{width=1\textwidth}
    \begin{tabular}{llllllllll}
    \hline
        \textbf{Architecture} & \textbf{Layer} & \textbf{Activation} & \textbf{Pooling} & \textbf{Dropout} & \textbf{Normalization} & \textbf{Image Dim.} & \textbf{Val. Acc.} & \textbf{Train Time (s)} & \textbf{Eval Time (s)} \\ \hline
        Rohrer68K & Conv2d & ReLU & MaxAbsPool2d & None & None & 32x32 & 0.6762 & 92.487 & 9.3 \\
        Rohrer68K & Conv2d & ReLU & MaxPool2d & None & None & 32x32 & 0.836 & 93.161 & 9.424 \\
        Rohrer68K & SCS & None & MaxAbsPool2d & None & None & 32x32 & 0.8032 & 129.316 & 8.253 \\
        Rohrer68K & SCS & None & MaxPool2d & None & None & 32x32 & 0.8103 & 128.442 & 7.998 \\
        Rohrer68K & SCS & ReLU & MaxAbsPool2d & None & None & 32x32 & 0.7594 & 128.578 & 8.031 \\
        Rohrer68K & SCS & ReLU & MaxPool2d & None & None & 32x32 & 0.8258 & 129.175 & 8.093 \\
        Rohrer68K & SharpenedSDP & None & MaxAbsPool2d & None & None & 32x32 & 0.8088 & 127.146 & 7.745 \\
        Rohrer68K & SharpenedSDP & None & MaxPool2d & None & None & 32x32 & 0.8261 & 127.014 & 7.747 \\
        Rohrer68K & SharpenedSDP & ReLU & MaxAbsPool2d & None & None & 32x32 & 0.7503 & 127.479 & 7.763 \\
        Rohrer68K & SharpenedSDP & ReLU & MaxPool2d & None & None & 32x32 & 0.8303 & 128.185 & 7.667 \\ \hline
    \end{tabular}
    \end{adjustbox}
    \label{table:rohrer68k_accuracies}
\end{table}

\begin{table}[H]
    \caption{Test accuracies of Rohrer100K model variants on $32 \times 32$ CIFAR-10 (new set of initial weights).\newline}
    \centering
    \begin{adjustbox}{width=1\textwidth}
    \begin{tabular}{llllllllll}
    \hline
        \textbf{Architecture} & \textbf{Layer} & \textbf{Activation} & \textbf{Pooling} & \textbf{Dropout} & \textbf{Normalization} & \textbf{Image Dim.} & \textbf{Val. Acc.} & \textbf{Train Time (s)} & \textbf{Eval Time (s)} \\ \hline
        Rohrer100K & Conv2d & ReLU & MaxAbsPool2d & None & None & 32x32 & 0.7307 & 93.024 & 9.34 \\
        Rohrer100K & Conv2d & ReLU & MaxPool2d & None & None & 32x32 & 0.8347 & 91.761 & 9.243 \\
        Rohrer100K & SCS & None & MaxAbsPool2d & None & None & 32x32 & 0.818 & 135.934 & 7.738 \\
        Rohrer100K & SCS & None & MaxPool2d & None & None & 32x32 & 0.8178 & 134.931 & 7.655 \\
        Rohrer100K & SCS & ReLU & MaxAbsPool2d & None & None & 32x32 & 0.7848 & 134.43 & 7.803 \\
        Rohrer100K & SCS & ReLU & MaxPool2d & None & None & 32x32 & 0.8252 & 132.427 & 7.702 \\
        Rohrer100K & SharpenedSDP & None & MaxAbsPool2d & None & None & 32x32 & 0.8206 & 130.617 & 7.374 \\
        Rohrer100K & SharpenedSDP & None & MaxPool2d & None & None & 32x32 & 0.8376 & 132.427 & 7.503 \\
        Rohrer100K & SharpenedSDP & ReLU & MaxAbsPool2d & None & None & 32x32 & 0.7605 & 132.941 & 7.481 \\
        Rohrer100K & SharpenedSDP & ReLU & MaxPool2d & None & None & 32x32 & 0.8243 & 132.174 & 7.342 \\ \hline
    \end{tabular}
    \end{adjustbox}
    \label{table:rohrer100k_accuracies}
\end{table}

\begin{table}[H]
    \caption{Test accuracies of Rohrer583K model variants on $32 \times 32$ CIFAR-10.\newline}
    \centering
    \begin{adjustbox}{width=1\textwidth}
    \begin{tabular}{llllllllll}
    \hline
        \textbf{Architecture} & \textbf{Layer} & \textbf{Activation} & \textbf{Pooling} & \textbf{Dropout} & \textbf{Normalization} & \textbf{Image Dim.} & \textbf{Val. Acc.} & \textbf{Train Time (s)} & \textbf{Eval Time (s)} \\ \hline
        Rohrer583K & Conv2d & ReLU & MaxAbsPool2d & None & None & 32x32 & 0.731 & 95.824 & 9.544 \\
        Rohrer583K & Conv2d & ReLU & MaxPool2d & None & None & 32x32 & 0.8587 & 94.684 & 9.404 \\
        Rohrer583K & SCS & None & MaxAbsPool2d & None & None & 32x32 & 0.8543 & 133.319 & 6.823 \\
        Rohrer583K & SCS & None & MaxPool2d & None & None & 32x32 & 0.8634 & 132.682 & 6.559 \\
        Rohrer583K & SCS & ReLU & MaxAbsPool2d & None & None & 32x32 & 0.8227 & 130.912 & 6.631 \\
        Rohrer583K & SCS & ReLU & MaxPool2d & None & None & 32x32 & 0.8611 & 133.041 & 6.69 \\
        Rohrer583K & SharpenedSDP & None & MaxAbsPool2d & None & None & 32x32 & 0.8319 & 131.443 & 6.271 \\
        Rohrer583K & SharpenedSDP & None & MaxPool2d & None & None & 32x32 & 0.8585 & 131.557 & 6.16 \\
        Rohrer583K & SharpenedSDP & ReLU & MaxAbsPool2d & None & None & 32x32 & 0.7679 & 131.32 & 6.32 \\
        Rohrer583K & SharpenedSDP & ReLU & MaxPool2d & None & None & 32x32 & 0.8471 & 131.278 & 6.297 \\ \hline
    \end{tabular}
    \end{adjustbox}
    \label{table:rohrer583k_accuracies}
\end{table}

\begin{table}[H]
    \caption{Test accuracies of ResNet18 model variants on $224 \times 224$ CIFAR-10.\newline}
    \centering
    \begin{adjustbox}{width=1\textwidth}
    \begin{tabular}{llllllllll}
    \hline
        \textbf{Architecture} & \textbf{Layer} & \textbf{Activation} & \textbf{Pooling} & \textbf{Dropout} & \textbf{Normalization} & \textbf{Image Dim.} & \textbf{Val. Acc.} & \textbf{Train Time (s)} & \textbf{Eval Time (s)} \\ \hline
        ResNet18 & Conv2d & ReLU & MaxAbsPool2d & None & BatchNorm2d & 224x224 & 0.9287 & 156.556 & 26.798 \\
        ResNet18 & Conv2d & ReLU & MaxAbsPool2d & None & None & 224x224 & 0.8835 & 160.957 & 23.954 \\
        ResNet18 & Conv2d & ReLU & MaxPool2d & None & BatchNorm2d & 224x224 & 0.9255 & 153.939 & 26.491 \\
        ResNet18 & Conv2d & ReLU & MaxPool2d & None & None & 224x224 & 0.8928 & 162.47 & 24.621 \\
        ResNet18 & SCS & None & MaxAbsPool2d & None & BatchNorm2d & 224x224 & 0.9164 & 192.729 & 27.624 \\
        ResNet18 & SCS & None & MaxAbsPool2d & None & None & 224x224 & 0.9006 & 191.534 & 27.098 \\
        ResNet18 & SCS & None & MaxPool2d & None & BatchNorm2d & 224x224 & 0.905 & 192.184 & 25.025 \\
        ResNet18 & SCS & None & MaxPool2d & None & None & 224x224 & 0.8972 & 188.621 & 26.262 \\
        ResNet18 & SCS & ReLU & MaxAbsPool2d & None & BatchNorm2d & 224x224 & 0.9219 & 193.221 & 27.219 \\
        ResNet18 & SCS & ReLU & MaxAbsPool2d & None & None & 224x224 & 0.8934 & 193.492 & 25.573 \\
        ResNet18 & SCS & ReLU & MaxPool2d & None & BatchNorm2d & 224x224 & 0.9191 & 192.479 & 24.804 \\
        ResNet18 & SCS & ReLU & MaxPool2d & None & None & 224x224 & 0.8853 & 193.119 & 22.343 \\ \hline
    \end{tabular}
    \end{adjustbox}
    \label{table:ResNet18_224_accuracies}
\end{table}

\subsection{Saliency Maps and Interpretability}
In this section, we only display saliency maps for model variants that achieved an overall accuracy of $60\%$ or greater. Models achieving lower accuracy than $60\%$ may not be representative of their layers' overall behaviors.
\begin{figure}[H]
\centering
  \subfloat[Saliency Maps]{
	\begin{minipage}[c][0.5\width]{
	   0.75\textwidth}
	   \centering
\includegraphics[width=1\textwidth]{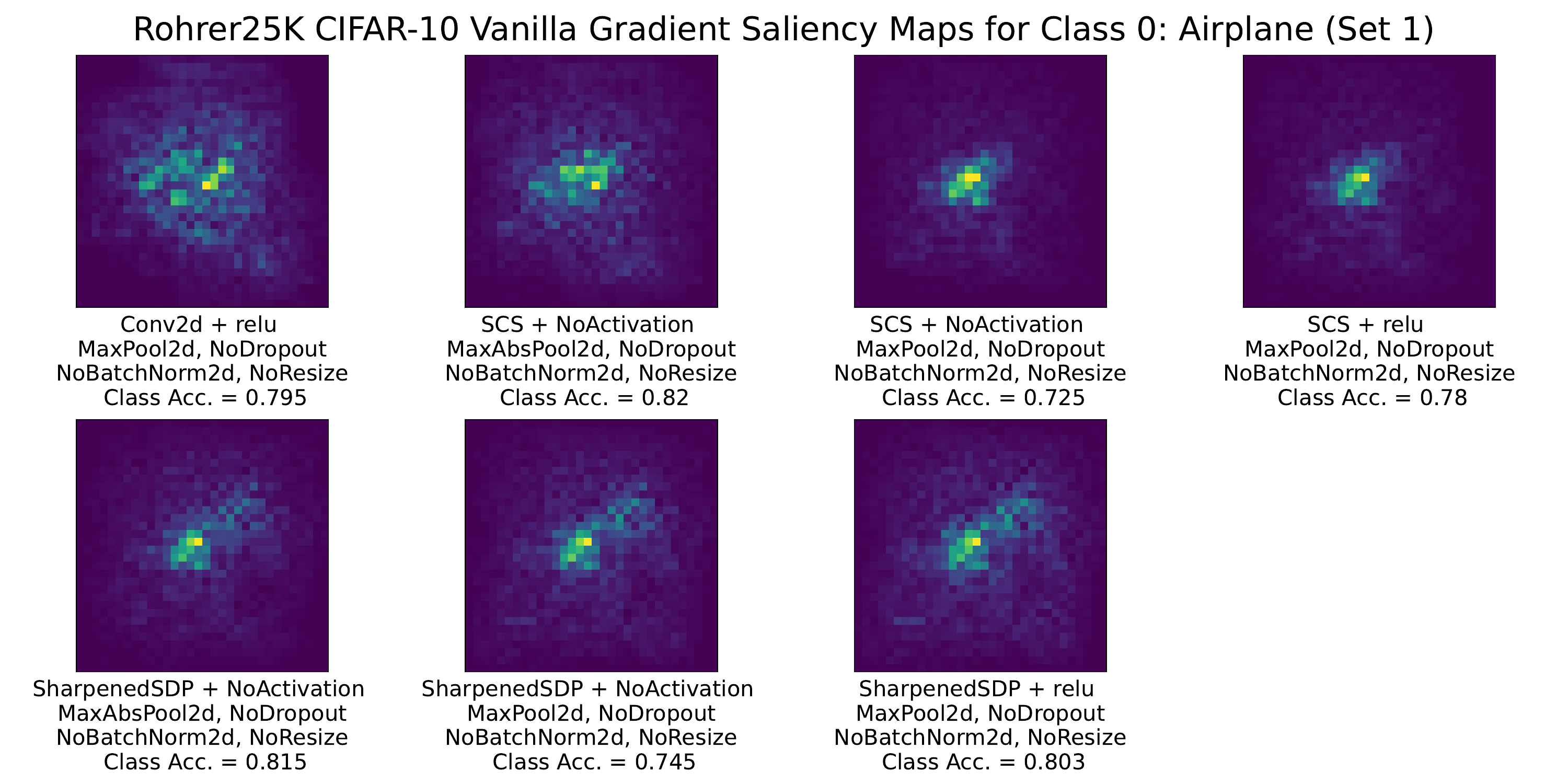}
	\end{minipage}}
  \subfloat[Original Image]{
	\begin{minipage}[c][1.2\width]{
	   0.18\textwidth}
	   \centering
	   \includegraphics[width=1\textwidth]{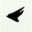}
	\end{minipage}}
\caption{Saliency maps of Rohrer25K variants on CIFAR-10. Note how both the SCS and SharpenedSDP saliency maps tend to be significantly more ``sparse" compared to the noisy convolutional map. This suggests that the $p$ exponentiation is what decisively determines saliency map ``sparsity."}
\label{fig:rohrer25k_saliency}
\end{figure}

\begin{figure}[H]
\centering
  \subfloat[Saliency Maps]{
	\begin{minipage}[c][0.5\width]{
	   0.75\textwidth}
	   \centering
\includegraphics[width=1\textwidth]{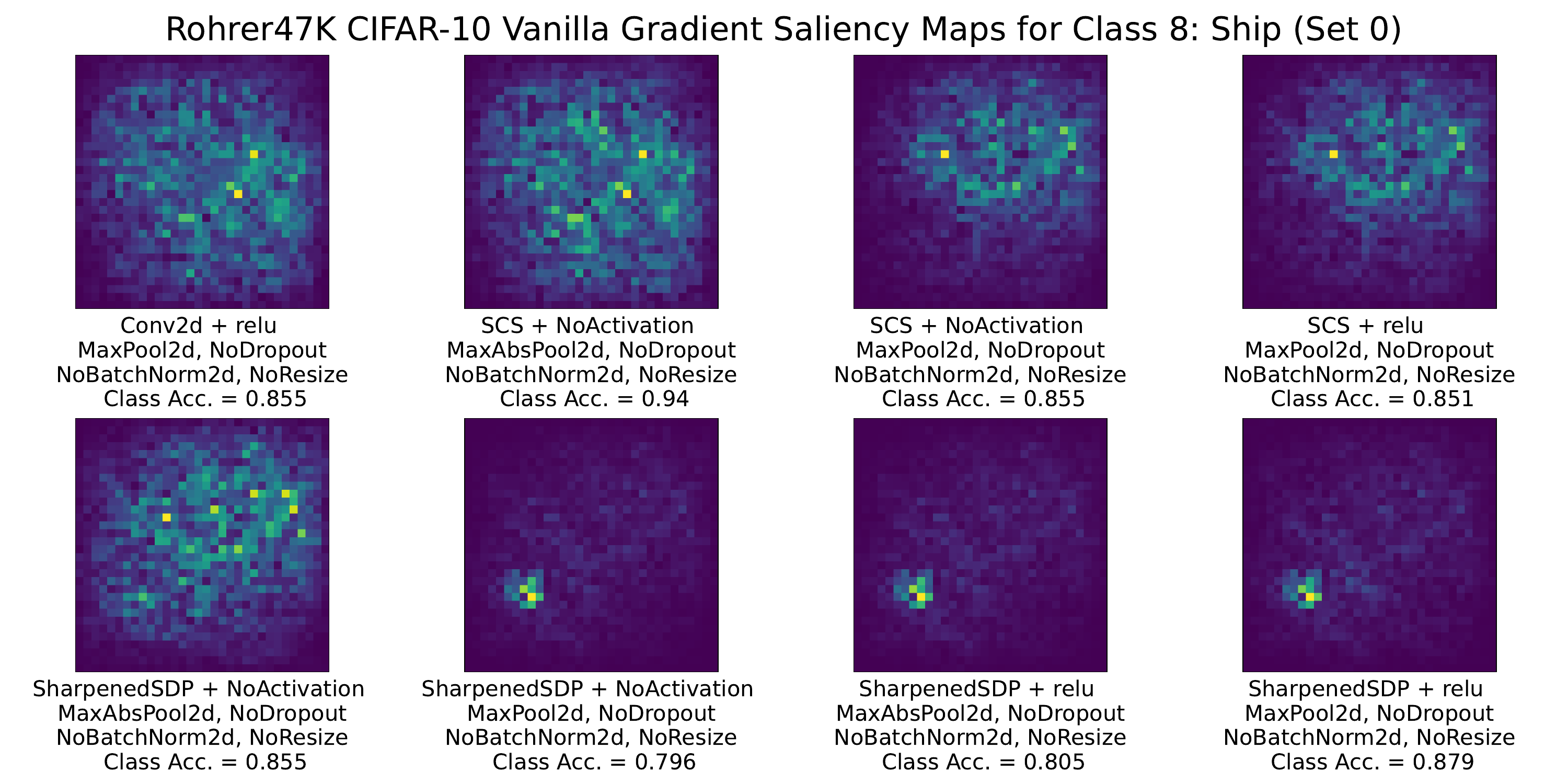}
	\end{minipage}}
  \subfloat[Original Image]{
	\begin{minipage}[c][1.2\width]{
	   0.18\textwidth}
	   \centering
	   \includegraphics[width=1\textwidth]{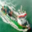}
	\end{minipage}}
\caption{Saliency maps of Rohrer47K variants on CIFAR-10. We find that SharpenedSDP may produce saliency maps that are noticeably sparser than SCS, but not immediately interpretable. More thorough investigation is needed to determine what causes the SharpenedSDP variants to focus on the bottom left of the image of the ship presented. Regardless, it appears that the sharpening process is the decisive component here.}
\label{fig:rohrer47k_saliency}
\end{figure}

\begin{figure}[H]
\centering
  \subfloat[Saliency Maps]{
	\begin{minipage}[c][0.6\width]{
	   0.75\textwidth}
	   \centering
\includegraphics[width=1\textwidth]{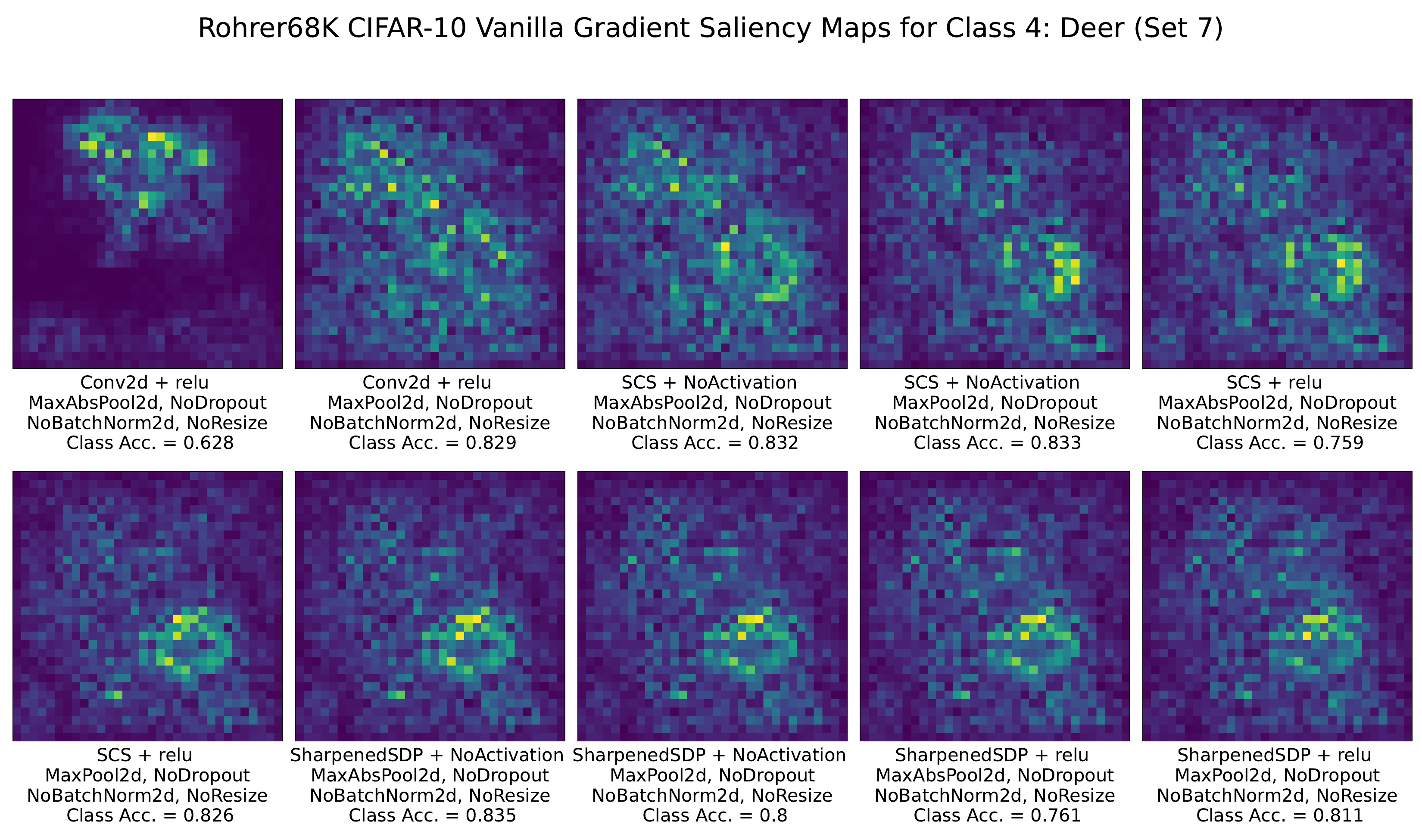}
	\end{minipage}}
  \subfloat[Original Image]{
	\begin{minipage}[c][1.2\width]{
	   0.18\textwidth}
	   \centering
	   \includegraphics[width=1\textwidth]{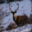}
	\end{minipage}}
\caption{Saliency maps of Rohrer68K variants on CIFAR-10. It appears that all of the model variants with sharpening (SCS and SharpenedSDP) seem to capture the posterior of the deer and arguably some antlers (specifically, SCS). The convolutional variant does not appear to capture any human-interpretable features in the deer image. We note how some of the SCS and SharpenedSDP saliency maps are almost indistinguishable. It is possible that exponentiation by $p$ serves to emphasize the signal.}
\label{fig:rohrer68k_saliency}
\end{figure}

\begin{figure}[H]
\centering
  \subfloat[Saliency Maps]{
	\begin{minipage}[c][0.6\width]{
	   0.75\textwidth}
	   \centering
\includegraphics[width=1\textwidth]{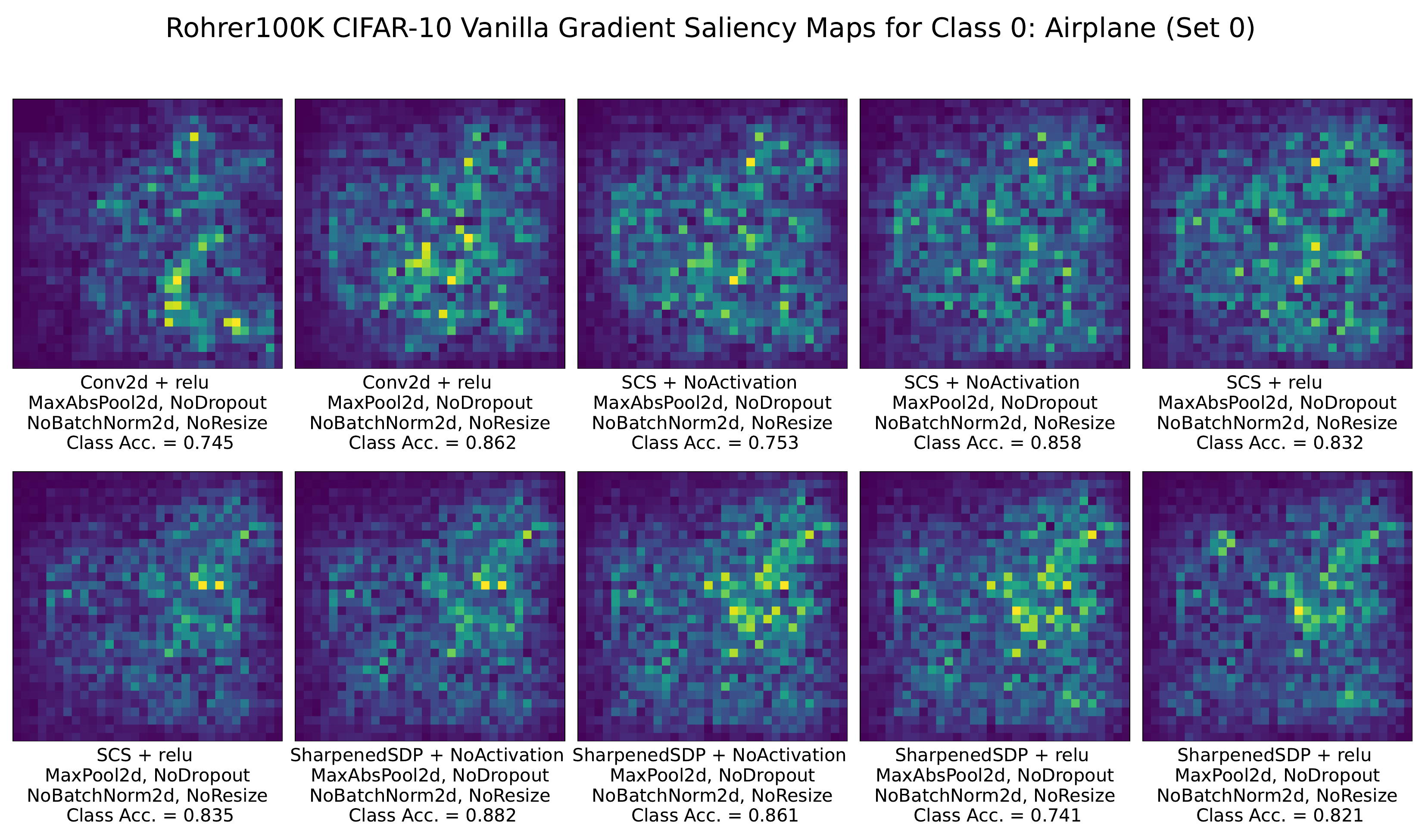}
	\end{minipage}}
  \subfloat[Original Image]{
	\begin{minipage}[c][1.2\width]{
	   0.18\textwidth}
	   \centering
	   \includegraphics[width=1\textwidth]{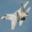}
	\end{minipage}}
\caption{Saliency maps of Rohrer100K variants on CIFAR-10. We see that 3 out of the 5 SharpenedSDP model variants clearly captured the right wing and stabilizer in the fighter jet image.  All convolutional and SCS variants did not appear to capture any meaningful, human-interpretable features in the image. Such a result (and many similar results encountered) suggest that SharpenedSDP itself may be a promising area of future exploration. Across examples, it appears that SharpenedSDP coupled with no activation tends to be a very promising combination.}
\label{fig:rohrer100k_saliency}
\end{figure}

\begin{figure}[H]
\centering
  \subfloat[Saliency Maps]{
	\begin{minipage}[c][0.6\width]{
	   0.75\textwidth}
	   \centering
\includegraphics[width=1\textwidth]{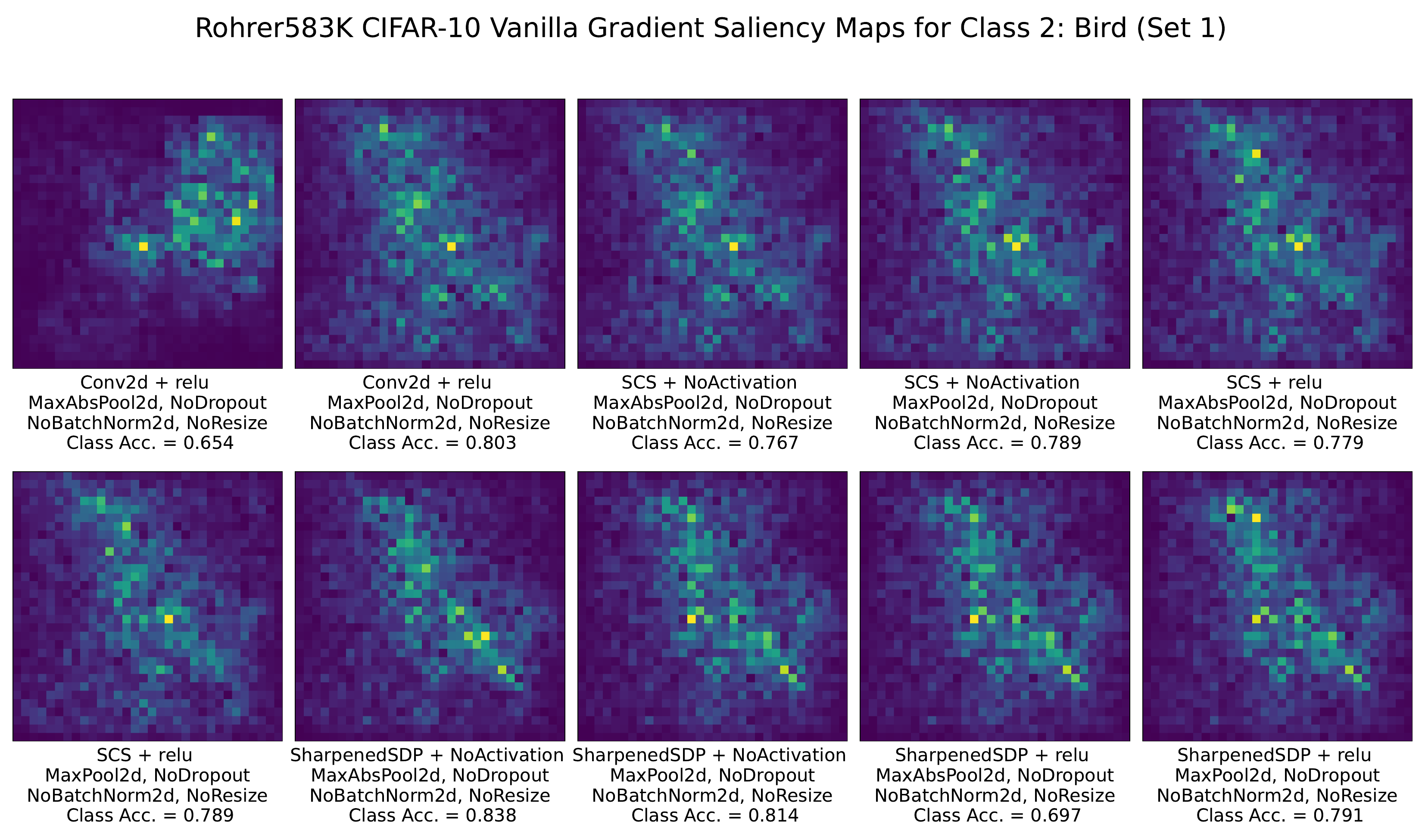}
	\end{minipage}}
  \subfloat[Original Image]{
	\begin{minipage}[c][1.2\width]{
	   0.18\textwidth}
	   \centering
	   \includegraphics[width=1\textwidth]{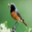}
	\end{minipage}}
\caption{Saliency maps of Rohrer583K variants on CIFAR-10. The purpose of this set of saliency maps is to show a case where all three layer-types produce saliency maps that look very similar to each other. This suggests that, at least theoretically, it is possible for various layer types to converge on the same identifying features. It is clear that all model variants (except the first) captured the silhouette of the bird.}
\label{fig:rohrer583k_saliency}
\end{figure}
\begin{figure}[H]
\centering
  \subfloat[Saliency Maps]{
	\begin{minipage}[c][0.99\width]{
	   0.75\textwidth}
	   \centering
\includegraphics[width=1\textwidth]{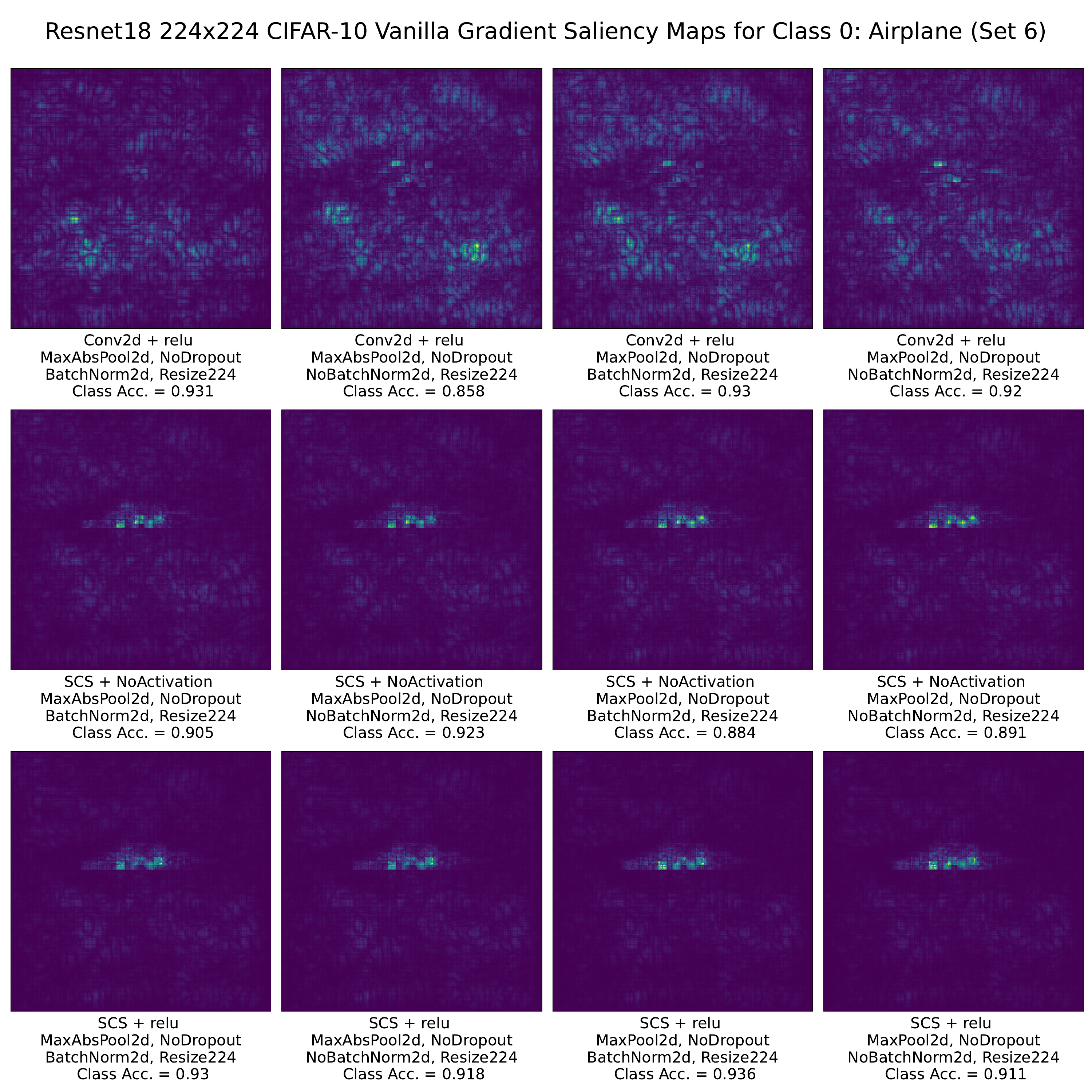}
	\end{minipage}}
  \subfloat[Original Image]{
	\begin{minipage}[c][1.2\width]{
	   0.18\textwidth}
	   \centering
	   \includegraphics[width=1\textwidth]{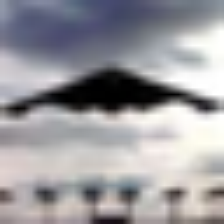}
	\end{minipage}}
\caption{Saliency maps of ResNet18 variants on $224 \times 224$ CIFAR-10 detecting a B-2 Spirit aircraft. Observe how the 4 convolution-based variants do not seem to produce human-interpretable saliency maps for the B-2 stealth bomber. However, all 8 SCS variants produced saliency maps that were extremely similar to each other, all capturing the jagged bottom edge of the aircraft. The SCS-based variants' saliency maps were also significantly sparser than that of the convolution-based variants. }
\label{fig:ResNet18_224_saliency}
\end{figure}

\subsection{Adversarial Robustness}
\begin{figure}[H]
\centering
  \subfloat[Rohrer25K]{
	\begin{minipage}[1.5\width]{
	   0.5\textwidth}
	   \centering
	   \input{}
	   \includegraphics[width=1\textwidth]{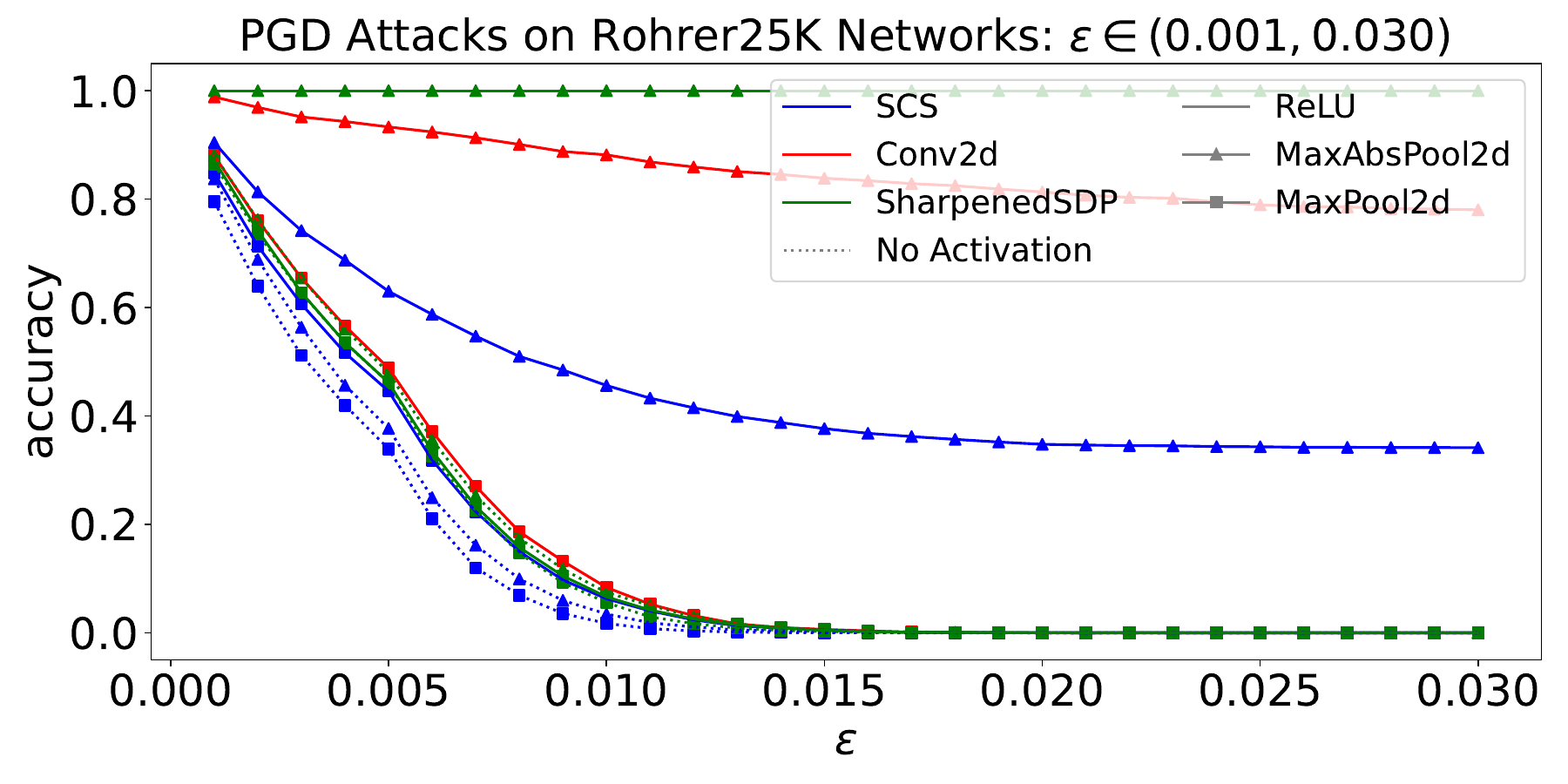}
	\end{minipage}}
  \subfloat[Rohrer47K]{
	\begin{minipage}[1.5\width]{
	   0.5\textwidth}
	   \centering
	   \includegraphics[width=1\textwidth]{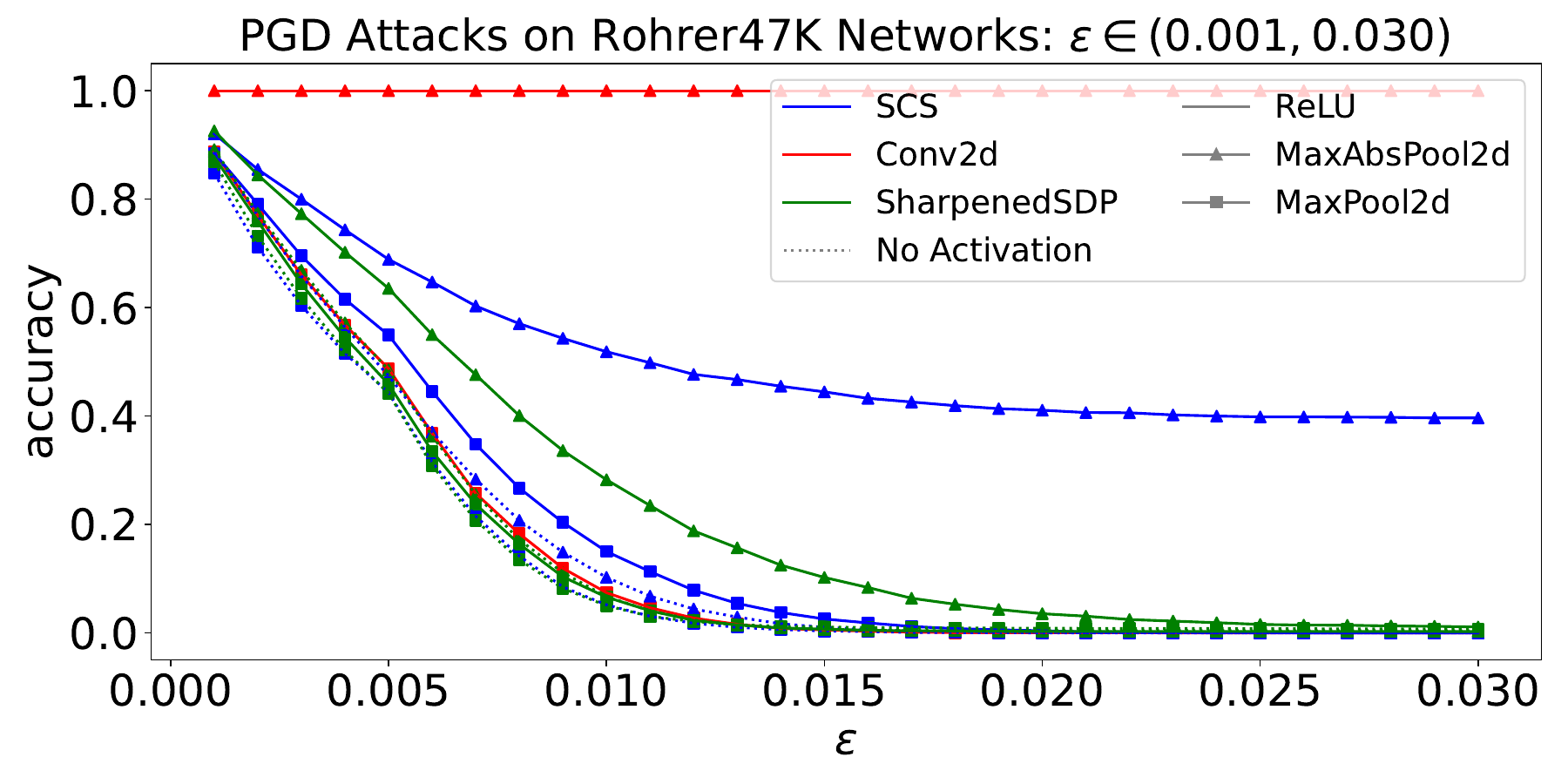}
	\end{minipage}}
 \hfill 	
  \subfloat[Rohrer68K]{
	\begin{minipage}[1.5\width]{
	   0.5\textwidth}
	   \centering
	   \includegraphics[width=1\textwidth]{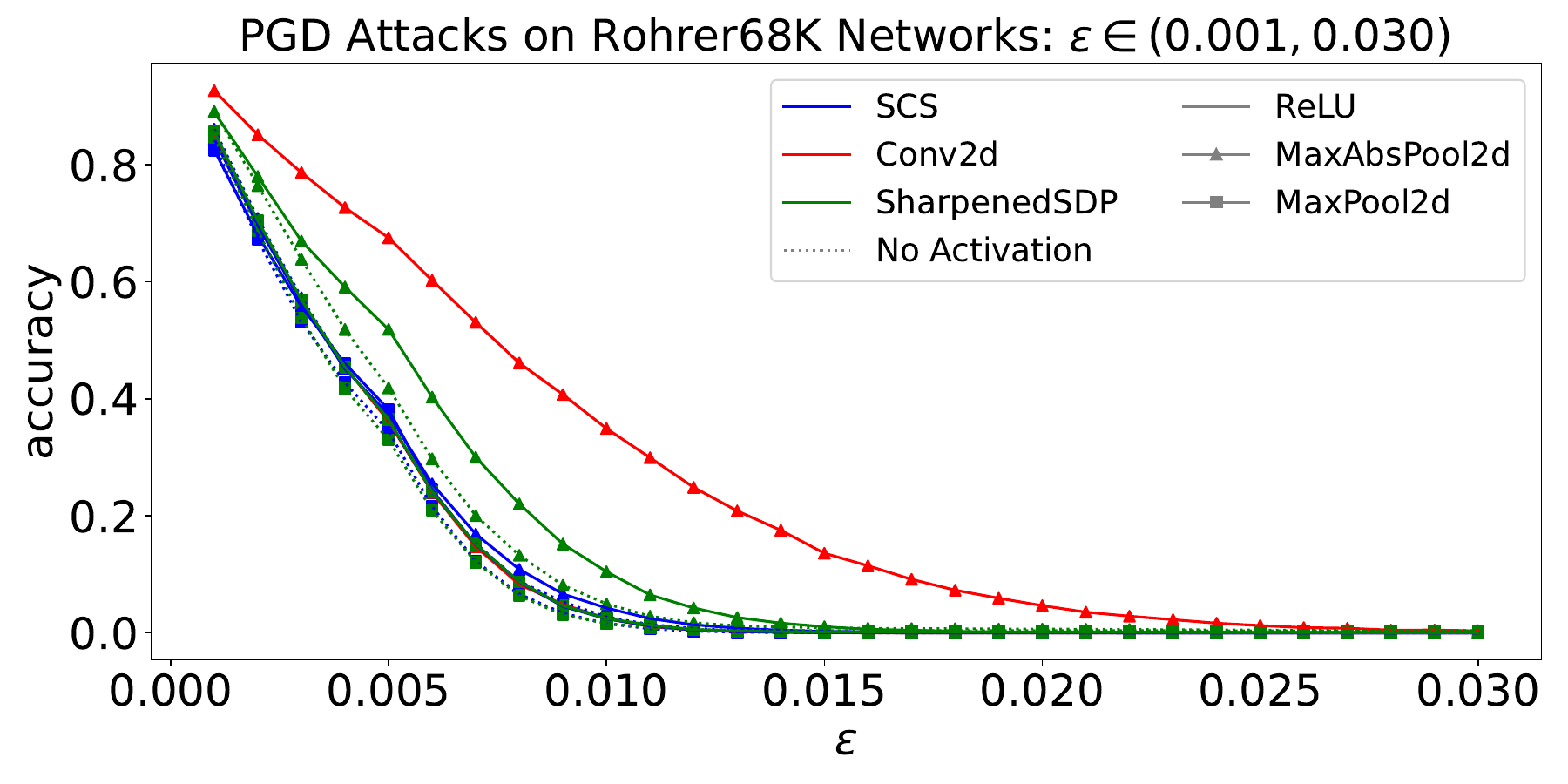}
	\end{minipage}}
  \subfloat[Rohrer583K]{
	\begin{minipage}[1.5\width]{
	   0.5\textwidth}
	   \centering
	   \includegraphics[width=1\textwidth]{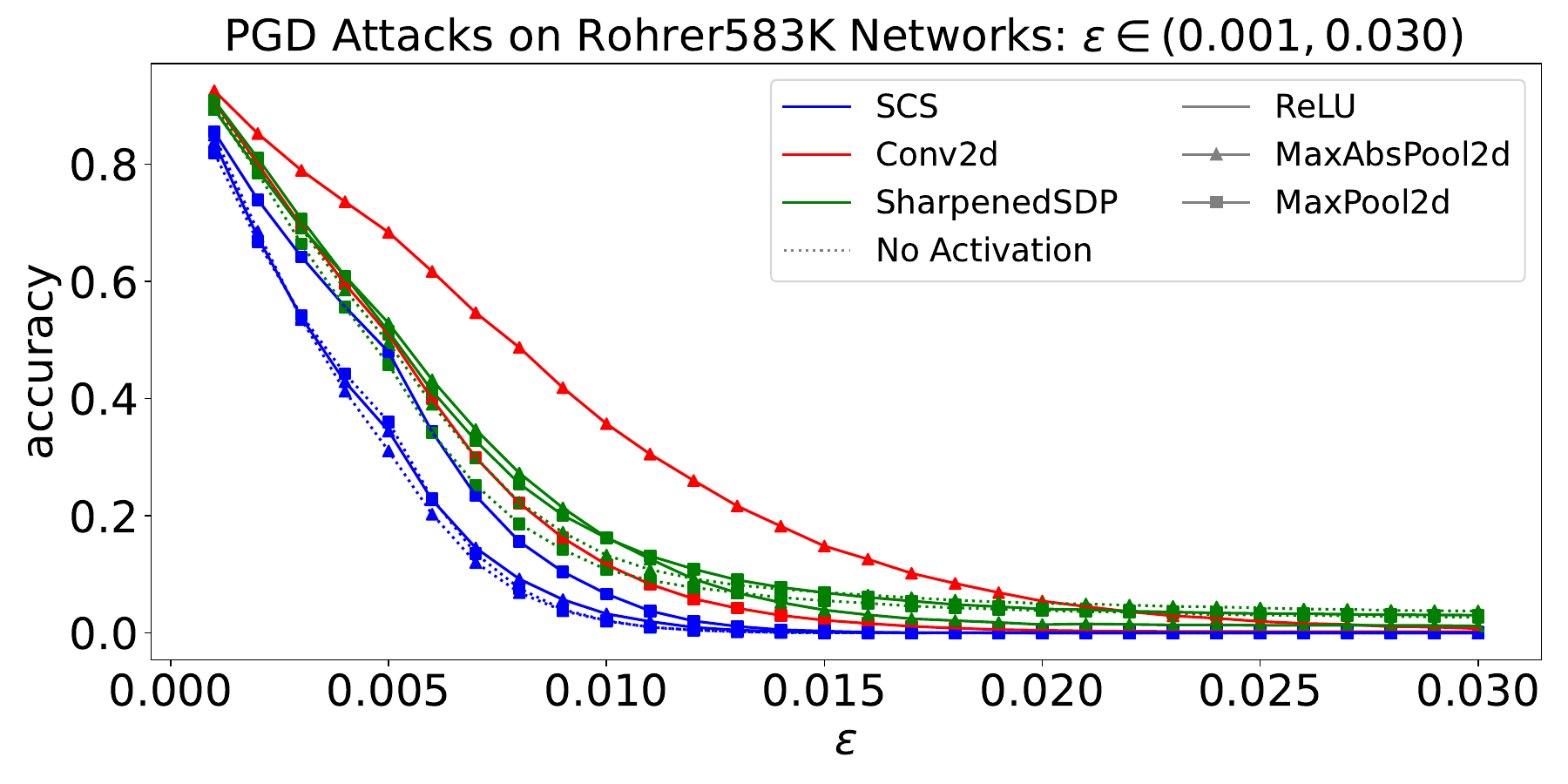}
	\end{minipage}}
\caption{PGD robustness of \texttt{scs-gallery} architecture variants on CIFAR-10 ($32 \times 32$).}
\label{fig:rohrer_family_PGD}
\end{figure}

\begin{figure}[H]
     \centering
     \includegraphics[scale = 0.30]{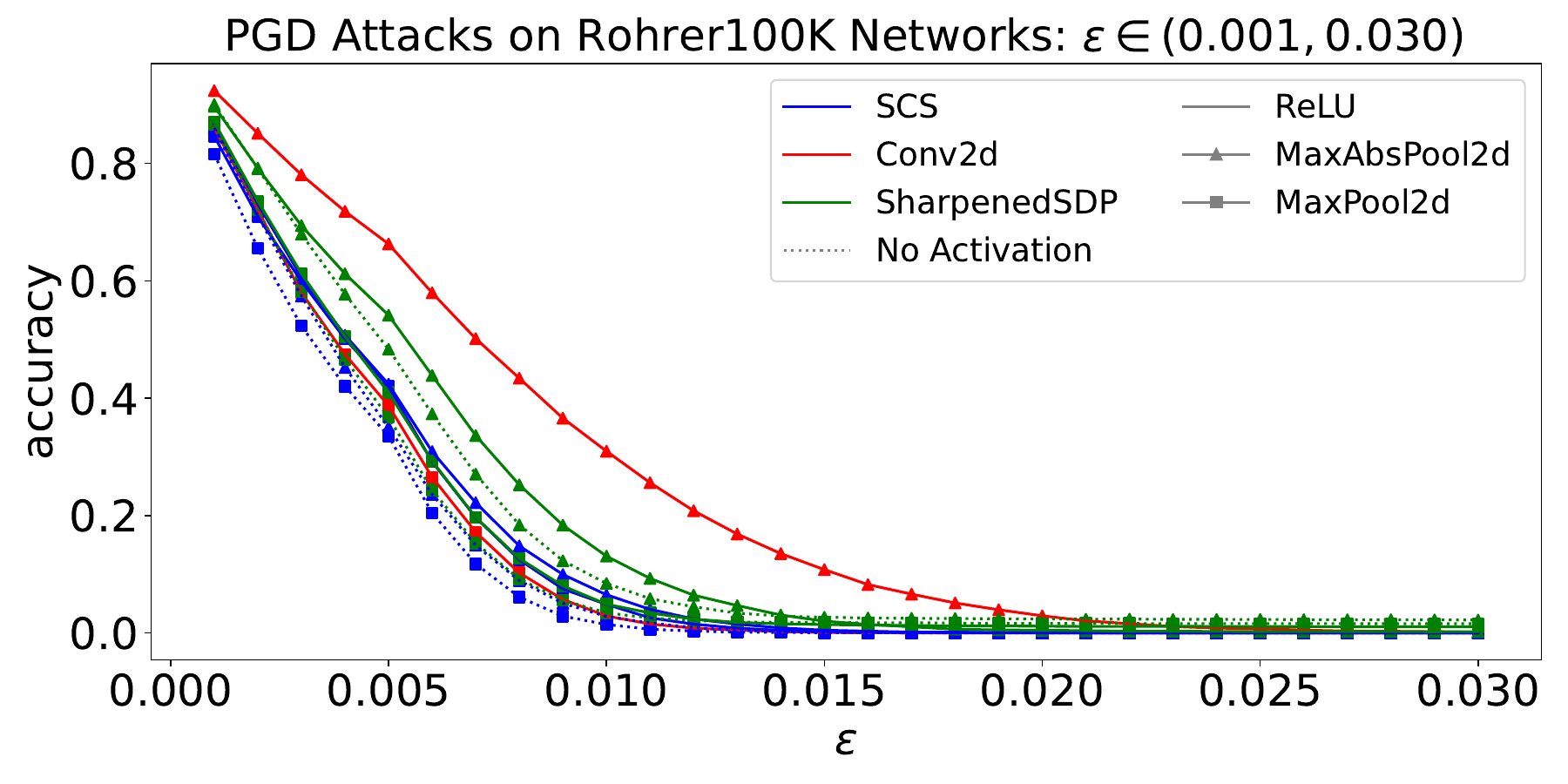}
     \caption{PGD robustness of Rohrer100K variants on CIFAR-10 ($32 \times 32$, additional testing).}
     \label{fig:rohrer100k_pgd_LATE}
\end{figure}

\begin{figure}[H]
\centering
  \subfloat[Without BatchNorm2d]{
	\begin{minipage}[c][0.6\width]{
	   0.48\textwidth}
	   \centering
	   \includegraphics[width=1\textwidth]{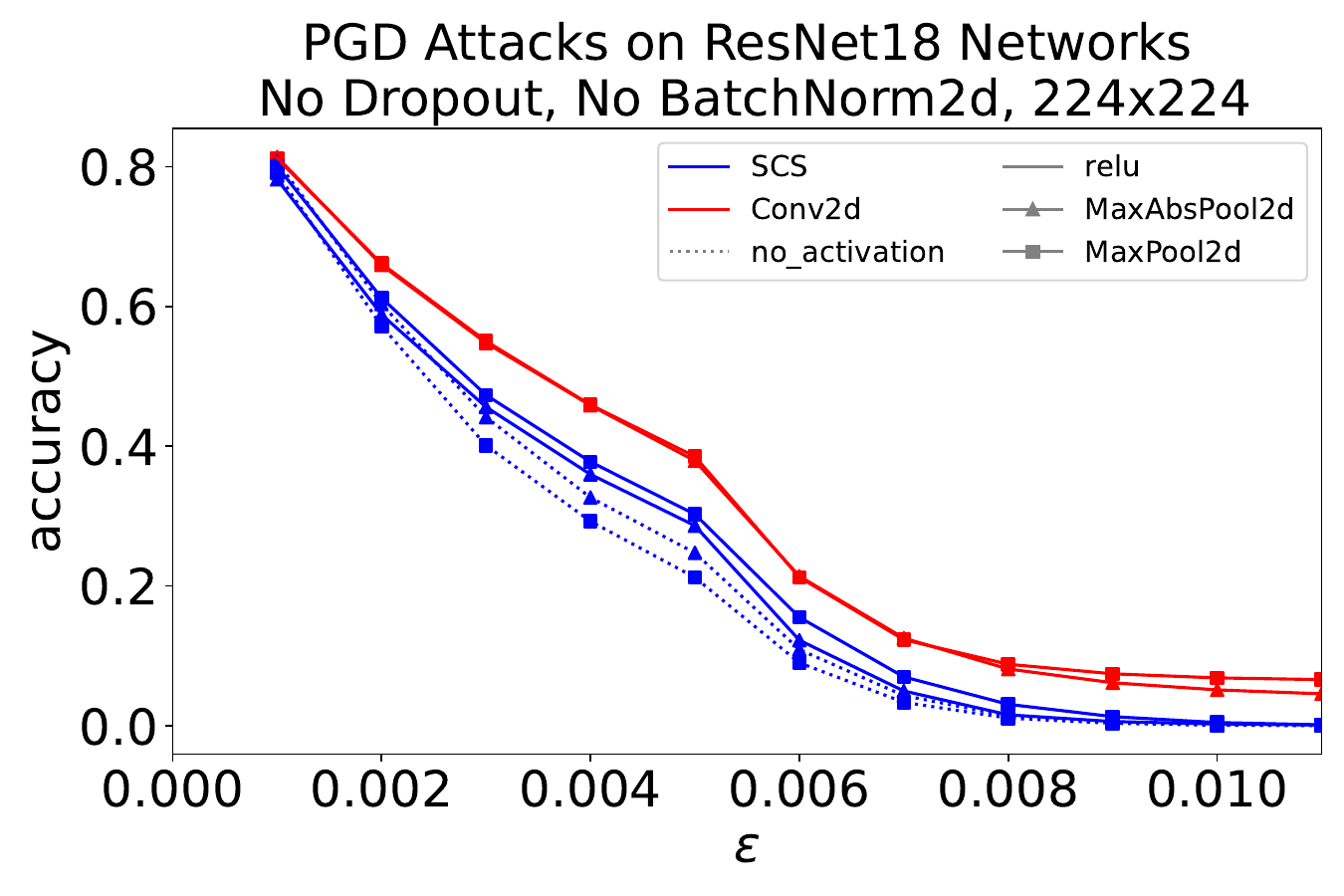}
	\end{minipage}}
  \subfloat[With BatchNorm2d]{
	\begin{minipage}[c][0.6\width]{
	   0.48\textwidth}
	   \centering
	   \includegraphics[width=1\textwidth]{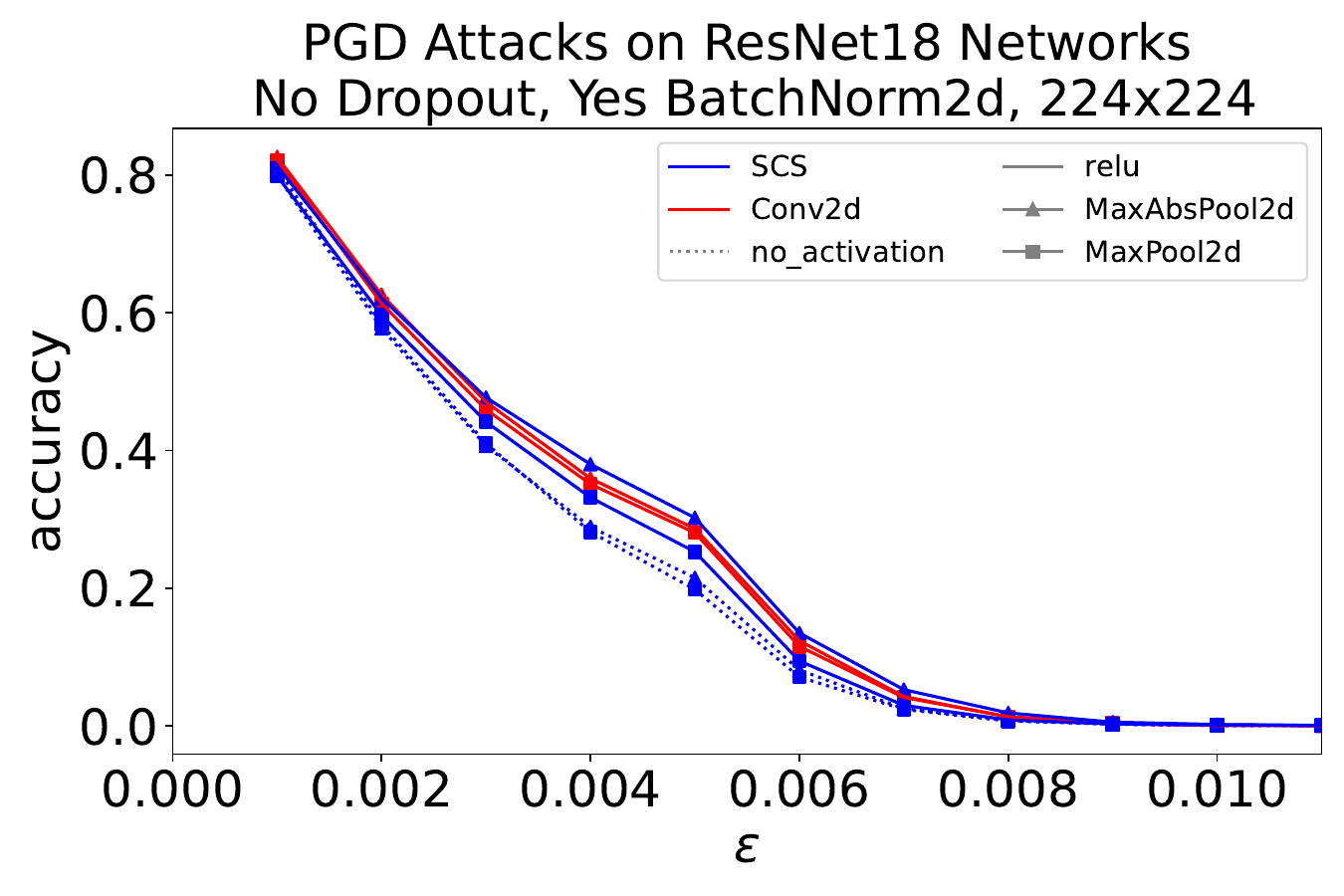}
	\end{minipage}}
\caption{PGD robustness of ResNet18 variants on $224 \times 224$ CIFAR-10.}
\label{fig:pgd_ResNet18_224}
\end{figure}
\end{document}